# Bayesian Optimization for Likelihood-Free Inference of Simulator-Based Statistical Models

**Michael U. Gutmann**                    MICHAEL.GUTMANN@HELSINKI.FI
*Helsinki Institute for Information Technology HIIT*
*Department of Mathematics and Statistics, University of Helsinki*
*Department of Information and Computer Science, Aalto University*

**Jukka Corander**                    JUKKA.CORANDER@HELSINKI.FI
*Helsinki Institute for Information Technology HIIT*
*Department of Mathematics and Statistics, University of Helsinki*

## Abstract

Our paper deals with inferring simulator-based statistical models given some observed data. A simulator-based model is a parametrized mechanism which specifies how data are generated. It is thus also referred to as generative model. We assume that only a finite number of parameters are of interest and allow the generative process to be very general; it may be a noisy nonlinear dynamical system with an unrestricted number of hidden variables. This weak assumption is useful for devising realistic models but it renders statistical inference very difficult. The main challenge is the intractability of the likelihood function. Several likelihood-free inference methods have been proposed which share the basic idea of identifying the parameters by finding values for which the discrepancy between simulated and observed data is small. A major obstacle to using these methods is their computational cost. The cost is largely due to the need to repeatedly simulate data sets and the lack of knowledge about how the parameters affect the discrepancy. We propose a strategy which combines probabilistic modeling of the discrepancy with optimization to facilitate likelihood-free inference. The strategy is implemented using Bayesian optimization and is shown to accelerate the inference through a reduction in the number of required simulations by several orders of magnitude.

## 1. Introduction

We consider the statistical inference of a finite number of parameters of interest $\boldsymbol{\theta} \in \mathbb{R}^d$ of a simulator-based statistical model for observed data $\mathbf{y}_o$ which consist of $n$ possibly dependent data points. A simulator-based statistical model is a parametrized stochastic data generating mechanism. Formally, it is a family of probability density functions (pdfs) $\{p_{\mathbf{y}|\boldsymbol{\theta}}\}_{\boldsymbol{\theta}}$ of unknown analytical form which allow for exact sampling of data $\mathbf{y}_{\boldsymbol{\theta}} \sim p_{\mathbf{y}|\boldsymbol{\theta}}$. In practical terms, it is a computer program which takes a value of $\boldsymbol{\theta}$ and a state of the random number generator as input and returns data $\mathbf{y}_{\boldsymbol{\theta}}$ as output. Simulator-based models are also called implicit models because the pdf of $\mathbf{y}_{\boldsymbol{\theta}}$ is not specified explicitly (Diggle and Gratton, 1984), or generative models because they specify how data are generated.

Simulator-based models are useful because they interface easily with models typically encountered in the natural sciences. In particular, hypotheses of how the observed data $\mathbf{y}_o$ were generated can be implemented without making excessive compromises in order to have an analytically tractable model pdf $p_{\mathbf{y}|\boldsymbol{\theta}}$.

Since the analytical form of $p_{\mathbf{y}|\boldsymbol{\theta}}$ is unknown, inference using the likelihood function $\mathcal{L}(\boldsymbol{\theta})$,

$$\mathcal{L}(\boldsymbol{\theta}) = p_{\mathbf{y}|\boldsymbol{\theta}}(\mathbf{y}_o|\boldsymbol{\theta}), \tag{1}$$

is not possible. The likelihood function is also not available for a large class of other statistical models which are known as unnormalized models. In these models, $p_{\mathbf{y}|\boldsymbol{\theta}}$ is only known up to a normalizing scaling factor (the partition function) which guarantees that $p_{\mathbf{y}|\boldsymbol{\theta}}$ is a valid pdf for all values of $\boldsymbol{\theta}$. Simulator-based models differ from unnormalized models in that not only is the scaling factor unknown but also the shape of $p_{\mathbf{y}|\boldsymbol{\theta}}$. Likelihood-free inference methods developed for unnormalized models (for example Hinton, 2002; Hyvärinen, 2005; Pihlaja et al., 2010; Gutmann and Hirayama, 2011; Gutmann and Hyvärinen, 2012) are thus not applicable to simulator-based models.

For simulator-based models, likelihood-free inference methods have emerged in multiple disciplines. "Indirect inference" originated in economics (Gouriéroux et al., 1993), "approximate Bayesian computation" (ABC) in genetics (Beaumont et al., 2002; Marjoram et al., 2003; Sisson et al., 2007), or the "synthetic likelihood" approach in ecology (Wood, 2010), for an overview, see, for example, the review by Hartig et al. (2011). The different methods share the basic idea to identify the model parameters by finding values which yield simulated data that resemble the observed data.

The generality of simulator-based models comes with the expense of two major difficulties in the inference. One difficulty is the assessment of the discrepancy between the observed and simulated data (Joyce and Marjoram, 2008; Wegmann et al., 2009; Nunes and Balding, 2010; Fearnhead and Prangle, 2012; Aeschbacher et al., 2012; Gutmann et al., 2014). The other difficulty is that the inference methods tend to be slow due to the need to simulate a large collection of data sets and due to the lack of knowledge about the relation between the model parameters and the corresponding discrepancies.

In this paper, we address the computational difficulty of the likelihood-free inference methods. We propose a strategy which combines probabilistic modeling of the discrepancies with optimization to facilitate likelihood-free inference. The strategy is implemented using Bayesian optimization (see, for example, Brochu et al., 2010). We show that using Bayesian optimization in likelihood-free inference (BOLFI) can reduce the number of required simulations by several orders of magnitude, which accelerates the inference substantially.[1]

The rest of the paper is organized as follows: In Section 2, we present examples of simulator-based statistical models to help clarify their properties. In Section 3, we provide a unified review of existing inference methods for simulator-based models, and use the examples to point out computational issues. The computational difficulties are summarized in Section 4, and a framework to address them is outlined in Section 5. Section 6 implements the framework using Bayesian optimization. Applications of the developed methodology are given in Section 7, and Section 8 concludes the paper.

---

1. Preliminary results were presented at "Approximate Bayesian Computation in Rome", 2013, and MCMCSki IV, 2014, as a poster "Bayesian optimization for efficient likelihood-free inference", and at the NIPS workshop "ABC in Montreal", 2014, as part of an oral presentation.

## 2. Examples of Simulator-Based Statistical Models

We present here three examples of simulator-based statistical models. The first example is an artificial one, but useful because it allows us to illustrate the central concepts. The other two are examples from real data analysis with intractable models (Wood, 2010; Numminen et al., 2013). The examples will be used throughout the paper and the model details can be looked up here when needed.

**Example 1** (Normal distribution). A standard way to sample data $\mathbf{y}_\theta = (y_\theta^{(1)}, \dots, y_\theta^{(n)})$ from a normal distribution with mean $\theta$ and variance one is to sample $n$ standard normal random variables $\boldsymbol{\omega} = (\omega^{(1)}, \dots, \omega^{(n)})$ and to add $\theta$ to the obtained samples,

$$\mathbf{y}_\theta = \theta + \boldsymbol{\omega}, \qquad\qquad \boldsymbol{\omega} \sim \mathcal{N}(0, \mathbf{I}_n). \qquad (2)$$

The symbol $\mathcal{N}(0, \mathbf{I}_n)$ denotes a $n$-variate normal distribution with mean zero and identity covariance matrix. After sampling of the random quantities $\boldsymbol{\omega}$, the observed data $\mathbf{y}_\theta$ are a deterministic transformation of $\boldsymbol{\omega}$ and the parameter $\theta$. For more general simulators, the same principle applies. In particular, the data $\mathbf{y}_{\boldsymbol{\theta}}$ are a deterministic transformation of $\boldsymbol{\theta}$ if the random quantities are kept fixed, for example by fixing the seed of the random number generator. ▲

**Example 2** (Ricker model). In this example, the simulator consists of a latent stochastic time series and an observation model. The latent time series is a stochastic version of the Ricker map which is a classical model in ecology (Ricker, 1954). The stochastic version can be described as a nonlinear autoregressive model,

$$\log N^{(t)} = \log r + \log N^{(t-1)} - N^{(t-1)} + \sigma e^{(t)}, \qquad t = 1, \dots, n, \qquad N^{(0)} = 0, \quad (3)$$

where $N^{(t)}$ is the size of some animal population at time $t$ and the $e^{(t)}$ are independent standard normal random variables. The latent time series has two parameters: $\log r$ which is related to the log growth rate and $\sigma$ for the standard deviation of the innovations. A Poisson observation model is assumed, such that given $N^{(t)}$, $y_{\boldsymbol{\theta}}^{(t)}$ is drawn from a Poisson distribution with mean $\varphi N^{(t)}$,

$$y_{\boldsymbol{\theta}}^{(t)} | N^{(t)}, \varphi \sim \text{Poisson}(\varphi N^{(t)}), \qquad\qquad (4)$$

where $\varphi$ is a scaling parameter. The model is thus in total parametrized by $\boldsymbol{\theta} = (\log r, \sigma, \varphi)$. Figure 1(a) shows example data generated from the model. Inference of $\boldsymbol{\theta}$ is difficult because the $N^{(t)}$ are not directly observed and because of the strong nonlinearity in the autoregressive model. Wood (2010) used this example to illustrate his "synthetic likelihood" approach to inference. ▲

**Example 3** (Bacterial infections in day care centers). The data generating process is here defined via a latent continuous-time Markov chain and an observation model. The model was developed by Numminen et al. (2013) to infer the transmission dynamics of bacterial infections in day care centers.

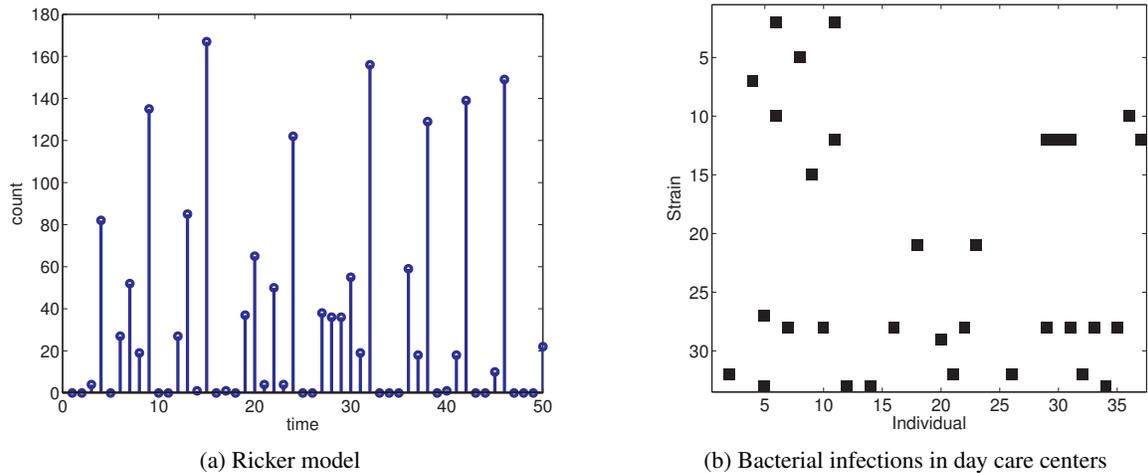

(a) Ricker model          (b) Bacterial infections in day care centers

Figure 1: Examples of simulator-based statistical models. (a) Data generated from the Ricker model in Example 2 with $n = 50$ and $\theta_o = (\log r_o, \sigma_o, \varphi_o) = (3.8, 0.3, 10)$. (b) Data generated from the model in Example 3 on bacterial infections in day care centers. There are 33 different strains of the bacterium in circulation and $M_{\mathrm{DCC}} = 36$ of the 53 attendees of the day care center were sampled (Numminen et al., 2013). Each black square indicates a sampled attendee who is infected with a particular strain. The data were generated with $\theta_o = (\beta_o, \Lambda_o, \theta_o) = (3.6, 0.6, 0.1)$.

The variables of the latent Markov chain are the binary indicator variables $I_{is}^t$ which specify whether attendee $i$ of a day care center is infected with the bacterial strain $s$ at time $t$ ($I_{is}^t = 1$), or not ($I_{is}^t = 0$). Starting with zero infected individuals, $I_{is}^0 = 0$ for all $i$ and $s$, the states evolve in a stochastic manner according to the rate equations

$$P(I_{is}^{t+h} = 0 | I_{is}^t = 1) = h + o(h), \tag{5}$$

$$P(I_{is}^{t+h} = 1 | I_{is'}^t = 0 \ \forall s') = R_s(t)h + o(h), \tag{6}$$

$$P(I_{is}^{t+h} = 1 | I_{is}^t = 0, \ \exists s' : I_{is'}^t = 1) = \theta R_s(t)h + o(h), \tag{7}$$

where $h$ is a small time interval and $R_s(t)$ the rate of infection with strain $s$ at time $t$. The three equations model the probability to clear a strain $s$ during time $t$ and $t + h$ (Equation 5), the probability to be infected with a strain $s$ if not colonized by other strains (Equation 6), and the probability to be infected if colonized with other strains (Equation 7). The rate of infection is a weighted combination of the probability $P_s$ for an infection happening outside the day care center and the probability $E_s(t)$ for an infection from within,

$$R_s(t) = \beta E_s(t) + \Lambda P_s. \tag{8}$$

We refer the reader to the original publication by Numminen et al. (2013) for more details and the expression for $E_s(t)$. The observation model was random sampling of $M_{\mathrm{DCC}}$ individuals without replacement from all the individuals attending a day care center at some sufficiently large random time (endemic situation). The model has three parameters $\boldsymbol{\theta} = (\beta, \Lambda, \theta)$: the internal infection parameter $\beta$, the external infection parameter $\Lambda$, and the co-infection parameter $\theta$. Figure 1(b) shows an example of data generated from the model.

Numminen et al. (2013) applied the model to data on colonizations with the bacterium *Streptococcus pneumoniae*. The observed data $\mathbf{y}_o$ were the states of the sampled attendees of 29 day care centers, that is, 29 binary matrices as in Figure 1(b) but with varying numbers of sampled attendees per day care center. Inference of the parameters is difficult because the data are a snapshot of the state of some of the attendees at a single time point only. Since the process evolves in continuous-time, the modeled system involves infinitely many correlated unobserved variables. ▲

## 3. Inference Methods for Simulator-Based Statistical Models

This section organizes the foundations and the previous work. We first point out properties common to all inference methods for simulator-based models, one being the general manner of constructing approximate likelihood functions. We then explain parametric and nonparametric approximations of the likelihood and discuss the relation between the two approaches. This is followed by a summary of currently used posterior inference schemes.

### 3.1 General Properties of the Different Inference Methods

Inference of simulator-based statistical models is generally based on some measurement of discrepancy $\Delta_{\boldsymbol{\theta}}$ between the observed data $\mathbf{y}_o$ and data $\mathbf{y}_{\boldsymbol{\theta}}$ simulated with parameter value $\boldsymbol{\theta}$. The discrepancy is used to define an approximation $\hat{L}(\boldsymbol{\theta})$ of the likelihood $\mathcal{L}(\boldsymbol{\theta})$. The approximation happens on multiple levels.

On a statistical level, the approximation consists of reducing the observed data $\mathbf{y}_o$ to some features, or summary statistics $\Phi_o$ before performing inference. The purpose of the summary statistics is to reduce the dimensionality and to filter out information which is not deemed relevant for the inference of $\boldsymbol{\theta}$. That is, in this first approximation, the likelihood $\mathcal{L}(\boldsymbol{\theta})$ is replaced with $L(\boldsymbol{\theta})$,

$$L(\boldsymbol{\theta}) = p_{\Phi|\boldsymbol{\theta}}(\Phi_o|\boldsymbol{\theta}), \tag{9}$$

where $p_{\Phi|\boldsymbol{\theta}}$ is the pdf of the summary statistics. The function $L(\boldsymbol{\theta})$ is a valid likelihood function, but for the inference of $\boldsymbol{\theta}$ given $\Phi_o$, and not for the inference of $\boldsymbol{\theta}$ given $\mathbf{y}_o$, in contrast to $\mathcal{L}(\boldsymbol{\theta})$, unless the chosen summary statistics happened to be sufficient in the standard statistical sense.

The likelihood function $L(\boldsymbol{\theta})$, however, is also not known, because the pdf $p_{\Phi|\boldsymbol{\theta}}$ is of unknown analytical form, which is a property inherited from $p_{\mathbf{y}|\boldsymbol{\theta}}$. Thus, $L(\boldsymbol{\theta})$ needs to be approximated by some method. We denote practical approximations obtained with finite computational resources by $\hat{L}(\boldsymbol{\theta})$. Limiting approximations if infinitely many computational resources were available will be denoted by $\tilde{L}(\boldsymbol{\theta})$.

In the paper, we will encounter several methods to construct $\hat{L}(\boldsymbol{\theta})$. They all base the approximation on simulated summary statistics $\Phi_{\boldsymbol{\theta}}$, generated with parameter value $\boldsymbol{\theta}$. The simulation of summary statistics is generally done by simulating a data set $\mathbf{y}_{\boldsymbol{\theta}}$, followed by its reduction to summary statistics. Table 1 provides an overview of the different "likelihoods" appearing in the paper.

| Symbol | Meaning | Definition |
|--------|---------|------------|
| $\mathcal{L}$ | true likelihood based on observed data | Eq (1) |
| $L$ | true likelihood based on summary statistics | Eq (9) |
| $\tilde{L}$ | approximation of $L$ requiring infinite computing power | Sec 3.1 |
| $\tilde{L}_s(\tilde{\ell}_s)$ | parametric approx/synthetic (log) likelihood | Eq (13) |
| $\tilde{L}_\kappa$ | nonparametric approx with kernel $\kappa$ | Eq (22) |
| $\tilde{L}_u$ | nonparametric approx with uniform kernel | Eq (25) |
| $\hat{L}$ | computable approximation of $L$ | Sec 3.1 |
| $\hat{L}_s^N(\hat{\ell}_s^N)$ | parametric approx/synthetic (log) likelihood with sample averages | Eq (15) |
| $\hat{L}_\kappa^N$ | nonparametric approx with kernel $\kappa$ and sample averages | Eq (21) |
| $\hat{L}_u^N$ | nonparametric approx with uniform kernel and sample averages | Eq (24) |
| $\hat{L}_s^{(t)}(\hat{\ell}_s^{(t)})$ | parametric approx/synthetic (log) likelihood with regression | Sec 5.2 |
| $\hat{L}_\kappa^{(t)}$ | nonparametric approx with kernel $\kappa$ and regression | Sec 5.3 |
| $\hat{L}_u^{(t)}$ | nonparametric approx with uniform kernel and regression | Sec 5.3 |

Table 1: The main (approximate) likelihood functions appearing in the paper. The superscript "$N$" indicates that the sample average is computed using $N$ simulated data sets per model parameter $\boldsymbol{\theta}$. The superscript "$(t)$" indicates that regression is performed with a training set containing $t$ simulated data sets. The parametric approximations will be used together with the Gaussian and the Ricker model, the nonparametric approximations together with the Gaussian and the day care center model.

After construction of $\hat{L}$, inference can be performed in the usual manner by replacing $\mathcal{L}$ with $\hat{L}$. Approximate posterior inference can be performed via Markov chain Monte Carlo (MCMC) algorithms or via an importance sampling approach (see, for example, Robert and Casella, 2004). The posterior expectation of a function $g(\boldsymbol{\theta})$ given $\mathbf{y}_o$ can be computed via importance sampling with auxiliary pdf $q(\boldsymbol{\theta})$,

$$\mathrm{E}(g(\boldsymbol{\theta})|\mathbf{y}_o) \approx \sum_{m=1}^{M} g(\boldsymbol{\theta}^{(m)}) w^{(m)}, \quad w^{(m)} = \frac{\mathcal{L}(\boldsymbol{\theta}^{(m)}) \frac{p_{\boldsymbol{\theta}}(\boldsymbol{\theta}^{(m)})}{q(\boldsymbol{\theta}^{(m)})}}{\sum_{i=1}^{M} \mathcal{L}(\boldsymbol{\theta}^{(i)}) \frac{p_{\boldsymbol{\theta}}(\boldsymbol{\theta}^{(i)})}{q(\boldsymbol{\theta}^{(i)})}}, \quad \boldsymbol{\theta}^{(m)} \stackrel{i.i.d.}{\sim} q(\boldsymbol{\theta}), \quad (10)$$

where $p_{\boldsymbol{\theta}}$ denotes the prior pdf. This approach also yields an estimate of the posterior distribution via the "particles" $\boldsymbol{\theta}^{(m)}$ and the associated weights $w^{(m)}$. A computable version is obtained by replacing $\mathcal{L}$ with $\hat{L}$, giving $\mathrm{E}(g(\boldsymbol{\theta})|\mathbf{y}_o) \approx \mathrm{E}(g(\boldsymbol{\theta})|\Phi_o)$,

$$\mathrm{E}(g(\boldsymbol{\theta})|\Phi_o) \approx \sum_{m=1}^{M} g(\boldsymbol{\theta}^{(m)}) \hat{w}^{(m)}, \quad \hat{w}^{(m)} = \frac{\hat{L}(\boldsymbol{\theta}^{(m)}) \frac{p_{\boldsymbol{\theta}}(\boldsymbol{\theta}^{(m)})}{q(\boldsymbol{\theta}^{(m)})}}{\sum_{i=1}^{M} \hat{L}(\boldsymbol{\theta}^{(i)}) \frac{p_{\boldsymbol{\theta}}(\boldsymbol{\theta}^{(i)})}{q(\boldsymbol{\theta}^{(i)})}}, \quad \boldsymbol{\theta}^{(m)} \stackrel{i.i.d.}{\sim} q(\boldsymbol{\theta}). \quad (11)$$

There is some flexibility in the choice of the auxiliary pdf $q(\boldsymbol{\theta})$ in Equations (10) and (11) which enables iterative adaptive algorithms where the accepted $\boldsymbol{\theta}^{(m)}$ of one iteration are used to define the auxiliary distribution $q(\boldsymbol{\theta})$ of the next iteration (population or sequential Monte Carlo algorithms, Cappé et al., 2004; Del Moral et al., 2006).

### 3.2 Parametric Approximation of the Likelihood

The pdf $p_{\Phi|\boldsymbol{\theta}}$ of the summary statistics is of unknown analytical form but it may be reasonably assumed that it belongs to a certain parametric family. For instance, if $\Phi_{\boldsymbol{\theta}}$ is obtained via averaging, the central limit theorem suggests that the pdf may be well approximated by a Gaussian distribution if the number of samples $n$ is sufficiently large,

$$p_{\Phi|\boldsymbol{\theta}}(\boldsymbol{\phi}|\boldsymbol{\theta}) \approx \frac{1}{(2\pi)^{p/2}|\det \boldsymbol{\Sigma_{\theta}}|^{1/2}} \exp\left(-\frac{1}{2}(\boldsymbol{\phi} - \boldsymbol{\mu_{\theta}})^{\top}\boldsymbol{\Sigma_{\theta}}^{-1}(\boldsymbol{\phi} - \boldsymbol{\mu_{\theta}})\right), \tag{12}$$

where $p$ is the dimension of $\Phi_{\boldsymbol{\theta}}$. The corresponding likelihood function is $\tilde{L}_s = \exp(\tilde{\ell}_s)$,

$$\tilde{\ell}_s(\boldsymbol{\theta}) = -\frac{p}{2}\log(2\pi) - \frac{1}{2}\log|\det \boldsymbol{\Sigma_{\theta}}| - \frac{1}{2}(\Phi_o - \boldsymbol{\mu_{\theta}})^{\top}\boldsymbol{\Sigma_{\theta}}^{-1}(\Phi_o - \boldsymbol{\mu_{\theta}}), \tag{13}$$

which is an approximation of $L(\boldsymbol{\theta})$ unless the summary statistics are indeed Gaussian. The mean $\boldsymbol{\mu_{\theta}}$ and the covariance matrix $\boldsymbol{\Sigma_{\theta}}$ are generally not known. But the simulator can be used to estimate them via a sample average $\mathrm{E}^N$ over $N$ independently generated summary statistics,

$$\hat{\boldsymbol{\mu_{\theta}}} = \mathrm{E}^N\left[\Phi_{\boldsymbol{\theta}}\right] = \frac{1}{N}\sum_{i=1}^{N}\Phi_{\boldsymbol{\theta}}^{(i)}, \quad \Phi_{\boldsymbol{\theta}}^{(i)} \overset{i.i.d.}{\sim} p_{\Phi|\boldsymbol{\theta}}, \quad \hat{\boldsymbol{\Sigma}}_{\boldsymbol{\theta}} = \mathrm{E}^N\left[(\Phi_{\boldsymbol{\theta}} - \hat{\boldsymbol{\mu_{\theta}}})(\Phi_{\boldsymbol{\theta}} - \hat{\boldsymbol{\mu_{\theta}}})^{\top}\right]. \tag{14}$$

A computable estimate $\hat{L}_s^N$ of the likelihood function $L(\boldsymbol{\theta})$ is then given by $\hat{L}_s^N = \exp(\hat{\ell}_s^N)$,

$$\hat{\ell}_s^N(\boldsymbol{\theta}) = -\frac{p}{2}\log(2\pi) - \frac{1}{2}\log|\det \hat{\boldsymbol{\Sigma}}_{\boldsymbol{\theta}}| - \frac{1}{2}(\Phi_o - \hat{\boldsymbol{\mu_{\theta}}})^{\top}\hat{\boldsymbol{\Sigma}}_{\boldsymbol{\theta}}^{-1}(\Phi_o - \hat{\boldsymbol{\mu_{\theta}}}). \tag{15}$$

This approximation was named synthetic likelihood (Wood, 2010), hence our subscript "s".

Due to the approximation of the expectation with a sample average, $\hat{\ell}_s^N$ is a stochastic process (a random function). We illustrate this in Example 4 below. We there also show that the number of simulated summary statistics (data sets) $N$ is a trade-off parameter: The computational cost decreases as $N$ decreases but the variability of the estimate increases as a consequence. It further turns out that the sample curves of $\hat{\ell}_s^N$ may not be smooth for finite $N$ and that decreasing $N$ may worsen their roughness. We illustrate this in Example 5 using the Ricker model.

**Example 4** (Synthetic likelihood for the mean of a normal distribution)**.** The sample average is a sufficient statistic for the task of inferring the mean $\theta$ from a sample $\mathbf{y}_o = (y_o^{(1)}, \ldots, y_o^{(n)})$ of a normal distribution with assumed variance one. We thus reduce the observed and simulated data $\mathbf{y}_o$ and $\mathbf{y}_\theta$ to the empirical means $\Phi_o$ and $\Phi_\theta$, respectively,

$$\Phi_o = \frac{1}{n}\sum_{i=1}^{n} y_o^{(i)}, \qquad\qquad \Phi_\theta = \frac{1}{n}\sum_{i=1}^{n} y_\theta^{(i)}. \tag{16}$$

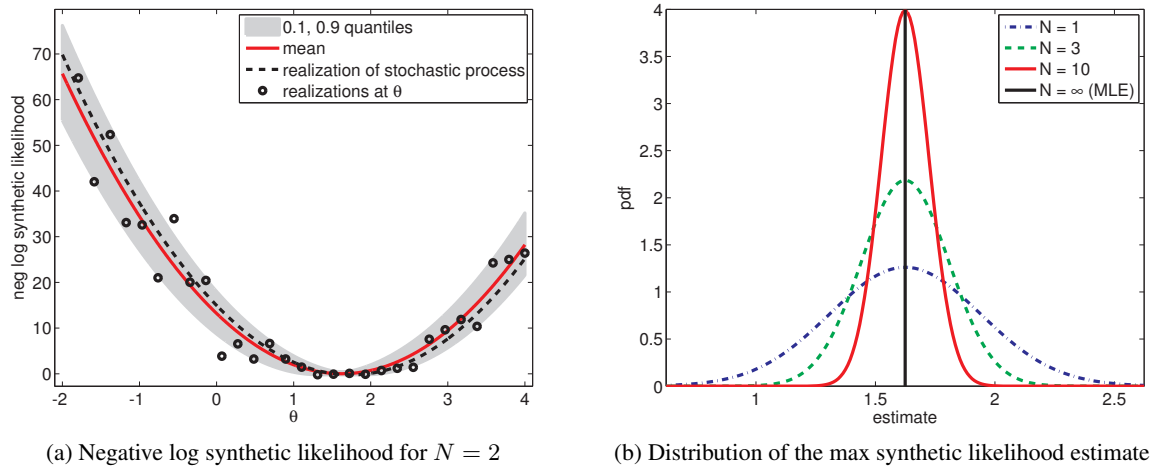

(a) Negative log synthetic likelihood for $N = 2$     (b) Distribution of the max synthetic likelihood estimate

Figure 2: Estimation of the mean of a Gaussian. (a) The figure shows the negative log synthetic likelihood $-\hat{\ell}_s^N$. It illustrates that $\hat{\ell}_s^N$ is a random function. (b) The randomness makes the estimate $\check{\theta} = \operatorname{argmax}_\theta \hat{\ell}_s^N(\theta)$ a random variable. Its variability increases as $N$ decreases.

In this special case, no information is lost with the reduction to the summary statistic, that is, $L(\theta) \propto \mathcal{L}(\theta)$. Furthermore, the distribution of the summary statistic $\Phi_\theta$ is here known, $\Phi_\theta \sim \mathcal{N}(\theta, 1/n)$ so that the Gaussian model assumption holds and $\bar{L}_s(\theta) = L(\theta)$.

Using for simplicity the true variance of $\Phi_\theta$, we have $\hat{\ell}_s^N(\theta) = -1/2 \log(2\pi/n) - n/2(\Phi_o - \hat{\mu}_\theta)^2$. Since $\hat{\mu}_\theta$ is an average of $N$ realizations of $\Phi_\theta$, $\hat{\mu}_\theta \sim \mathcal{N}(\theta, 1/(nN))$, and we can write $\hat{\ell}_s^N$ as a quadratic function subject to a random shift $g$,

$$\hat{\ell}_s^N(\theta) = -\frac{1}{2} \log\left(\frac{2\pi}{n}\right) - \frac{n}{2}(\Phi_o - \theta - g)^2, \qquad g \sim \mathcal{N}\left(0, \frac{1}{nN}\right). \qquad (17)$$

Each realization of $g$ yields a different mapping $\theta \mapsto \hat{\ell}_s^N$ which illustrates that the (log) synthetic likelihood is a random function. Figure 2(a) shows the 0.1 and 0.9 quantiles of $-\hat{\ell}_s^N$ for $N = 2$. The dashed curve visualizes $\theta \mapsto -\hat{\ell}_s^N$ for a fixed realization of $g$. The circles show values of $-\hat{\ell}_s^N(\theta)$ when $g$ is not kept fixed as $\theta$ changes. The results are for sample size $n = 10$.

The optimizer $\check{\theta}$ of each realization of $\hat{\ell}_s^N$ depends on $g$, $\check{\theta} = \Phi_o - g$. That is, $\check{\theta}$ is a random variable with distribution $\mathcal{N}(\Phi_o, 1/(Nn))$. In the limit of an infinite amount of available computational resources, that is $N \to \infty$, $g$ equals zero, and the distribution has a point-mass at $\hat{\theta}_{\mathrm{mle}} = \Phi_o$ which is indicated with the black vertical line in Figure 2(b). As $N$ decreases, variance is added to the point-estimate $\check{\theta}$. This added variability is due to the use of finite computational resources; it does not reflect uncertainty about $\theta$ due to the finite sample size $n$. The variability causes an inflation of the mean squared estimation error by a factor of $(1 + 1/N)$, $\mathrm{E}((\check{\theta} - \theta_o)^2) = 1/n(1 + 1/N)$. ▲

**Example 5** (Synthetic likelihood for the Ricker model). Wood (2010) used the synthetic likelihood to perform inference of the Ricker model and other simulator-based models with

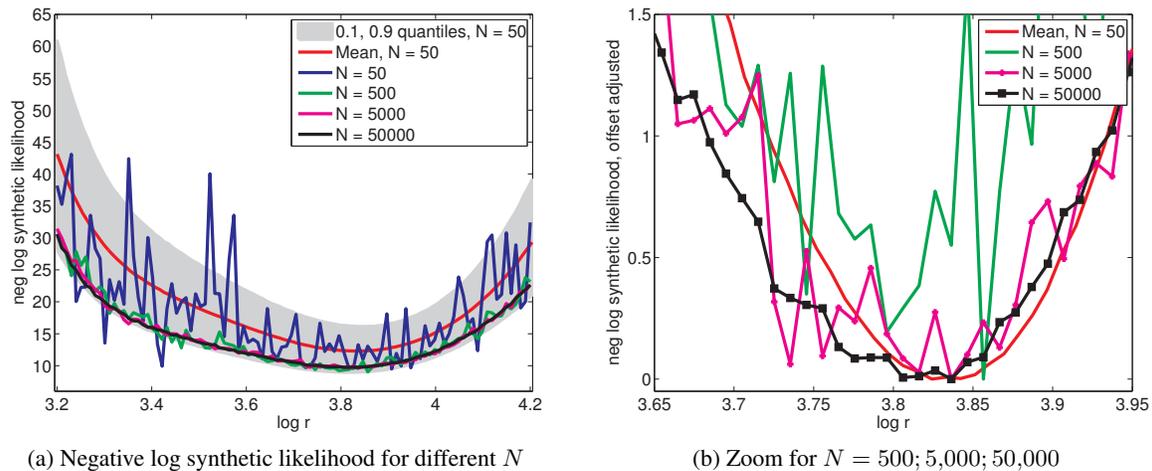

(a) Negative log synthetic likelihood for different $N$    (b) Zoom for $N = 500; 5,000; 50,000$

Figure 3: Using less computational resources may reduce the smoothness of the approximate likelihood function. The figures show the negative log synthetic likelihood $-\hat{\ell}_s^N$ for the Ricker model. Only the first parameter ($\log r$) was varied, the others were kept fixed at the data generating values. (a) The use of simulations makes the synthetic likelihood a stochastic process. Realizations of $-\hat{\ell}_s^N$ for different $N$ are shown together with the variability for $N = 50$. (b) The curves become more and more smooth as the number $N$ of simulated data sets increases even though the curve for $N = 50,000$ is still rugged. It is reasonable to assume though that the limit for $N \to \infty$ is smooth.

complex dynamics. Time series data $\mathbf{y}_{\boldsymbol{\theta}} = (y_{\boldsymbol{\theta}}^{(T_b+1)}, \ldots, y_{\boldsymbol{\theta}}^{(T_b+n)})$ from the Ricker model after some "burn-in" time $T_b$ were summarized in the form of the coefficients of the autocorrelation function and the coefficients of fitted nonlinear autoregressive models, thereby reducing the data to fourteen summary statistics $\Phi_{\boldsymbol{\theta}}$ (see the supplementary material of Wood, 2010, for their exact definition).

Figure 3 shows the negative log synthetic likelihood $-\hat{\ell}_s^N$ for the Ricker model as a function of the log growth rate $\log r$ for $\mathbf{y}_o$ in Figure 1(a). The parameters $\sigma$ and $\varphi$ were kept fixed at the values $\sigma_o = 0.3$ and $\varphi_o = 10$ which we used to generate $\mathbf{y}_o$ ($\log r^o$ was 3.8). The figures show that the realizations of the synthetic likelihood become less smooth as $N$ decreases.

The lack of smoothness makes the minimization of the different realizations of $-\hat{\ell}_s^N$ difficult. A grid-search is feasible for very large $N$ but this approach does not scale to higher dimensions. Gradient-based optimization is tricky because the functional form of $\hat{\ell}_s^N$ is unknown. Finite differences may not yield a reliable approximation of the gradient because of the lack of smoothness. Instead of optimizing a single realization of the objective, one could use an approximate stochastic gradient approach. That is, approximate gradients are computed with different random seeds at different values of the parameter. For small $N$, however, the gradients are unreliable so that the stepsize has to be very small, which makes the optimization rather costly again. To resolve the issue, we suggest a more efficient approach by combining probabilistic modeling with optimization.        ▲

### 3.3 Nonparametric Approximation of the Likelihood

An alternative to assuming a parametric model for the pdf $p_{\Phi|\theta}$ of the summary statistics is to approximate it by a kernel density estimate (Rosenblatt, 1956; Parzen, 1962; Mack and Rosenblatt, 1979; Wand and Jones, 1995),

$$p_{\Phi|\theta}(\phi|\theta) \approx \mathrm{E}^N\left[K(\phi, \Phi_\theta)\right],\tag{18}$$

where $K$ is a suitable kernel and $\mathrm{E}^N$ denotes empirical expectation as before,

$$\mathrm{E}^N\left[K(\phi, \Phi_\theta)\right] = \frac{1}{N}\sum_{i=1}^N K(\phi, \Phi_\theta^{(i)}), \qquad \Phi_\theta^{(i)} \overset{i.i.d.}{\sim} p_{\Phi|\theta}.\tag{19}$$

An approximation of the likelihood function $L(\theta)$ is given by $\hat{L}_K^N(\theta)$,

$$\hat{L}_K^N(\theta) = \mathrm{E}^N\left[K(\Phi_o, \Phi_\theta)\right].\tag{20}$$

We may re-write $K(\Phi_o, \Phi)$ in another form as $\kappa(\Delta_\theta)$ where $\Delta_\theta \geq 0$ depends on $\Phi_o$ and $\Phi_\theta$, and $\kappa$ is a univariate non-negative function not depending on $\theta$. The kernels $K$ are generally such that $\kappa$ has a maximum at zero (the maximum may be not unique though). Taking the empirical expectation in Equation (20) with respect to $\Delta_\theta$ instead of $\Phi_\theta$, we have $\hat{L}_K^N(\theta) = \hat{L}_\kappa^N(\theta)$,

$$\hat{L}_\kappa^N(\theta) = \mathrm{E}^N\left[\kappa\left(\Delta_\theta\right)\right].\tag{21}$$

As the number $N$ grows, $\hat{L}_\kappa^N$ converges to $\tilde{L}_\kappa$,

$$\tilde{L}_\kappa(\theta) = \mathrm{E}\left[\kappa(\Delta_\theta)\right],\tag{22}$$

which is $\hat{L}_\kappa^N$ where the empirical average $\mathrm{E}^N$ is replaced by the expectation $\mathrm{E}$. The limiting approximate likelihood $\tilde{L}_\kappa(\theta)$ does not not necessarily equal the likelihood $L(\theta) = p_{\Phi|\theta}(\Phi_o|\theta)$. For example, if $\kappa(\Delta_\theta)$ is obtained from a translation invariant kernel $K$, that is, $\kappa(\Delta_\theta) = K(\Phi_o - \Phi_\theta)$, $\tilde{L}_\kappa$ is the likelihood for a summary statistics whose pdf is obtained by convolving $p_{\Phi|\theta}$ with $K$.

For convex functions $\kappa$, Jensen's inequality yields a lower bound for $\hat{L}_\kappa^N$ and its logarithm,

$$\hat{L}_\kappa^N(\theta) \geq \kappa\left(\hat{J}^N(\theta)\right), \qquad \log \hat{L}_\kappa^N(\theta) \geq \log \kappa\left(\hat{J}^N(\theta)\right), \qquad \hat{J}^N(\theta) = \mathrm{E}^N\left[\Delta_\theta\right].\tag{23}$$

Since $\kappa$ is maximal at zero, the lower bound is maximized by minimizing the conditional empirical expectation $\hat{J}^N(\theta)$. The advantage of the lower bound is that it can be maximized irrespective of $\kappa$, which is often difficult to choose in practice.

A popular choice of $\kappa$ for likelihood-free inference is the uniform kernel $\kappa = \kappa_u$ which yields the approximate likelihood $\hat{L}_u^N$,

$$\kappa_u(u) = c\chi_{[0,h)}(u), \qquad \hat{L}_u^N(\theta) = c\,\mathrm{P}^N\left(\Delta_\theta < h\right),\tag{24}$$

where the indicator function $\chi_{[0,h)}(u)$ equals one if $u \in [0, h)$ and zero otherwise. The scaling parameter $c$ does not depend on $\theta$, and the positive scalar $h$ is the bandwidth of the kernel and acts as acceptance/rejection threshold. The approximate likelihood $\hat{L}_u^N$ is proportional to the empirical probability that the discrepancy is below the threshold. The limiting approximate likelihood is denoted by $\tilde{L}_u(\boldsymbol{\theta})$,

$$\tilde{L}_u(\boldsymbol{\theta}) = c \, \mathrm{P}\left(\Delta_{\boldsymbol{\theta}} < h\right). \tag{25}$$

The lower bound for convex $\kappa$ is not applicable but we can obtain an equivalent bound by Markov's inequality,

$$\hat{L}_u^N(\boldsymbol{\theta}) = c \left[1 - \mathrm{P}^N\left(\Delta_{\boldsymbol{\theta}}^u \geq h\right)\right] \geq c \left[1 - \frac{1}{h} \mathrm{E}^N\left[\Delta_{\boldsymbol{\theta}}\right]\right]. \tag{26}$$

The lower bound of the approximate likelihood can be maximized by minimizing $\hat{J}^N(\boldsymbol{\theta})$ as for convex $\kappa$.

We illustrate the approximation of the likelihood via $\hat{L}_u^N$ in Example 6 below. It is pointed out that good approximations are computationally very expensive because of the very small probability for $\Delta_{\boldsymbol{\theta}}$ to be below small thresholds $h$, or, in other words, because of the large rejection probability. We then use the model for bacterial infections in day care centers to show in Example 7 that the minimizer of $\hat{J}^N(\boldsymbol{\theta})$ can provide a good approximation of the maximizer of $\hat{L}_u^N(\boldsymbol{\theta})$. This is important because $\hat{J}^N$ does not require choosing the bandwidth $h$ or involve any rejections.

**Example 6** (Approximate likelihood for the mean of a Gaussian). For the inference of the mean of a Gaussian, we can use as discrepancy $\Delta_{\theta}$ the squared difference between the empirical mean of the observed and simulated data $\mathbf{y}_o$ and $\mathbf{y}_{\theta}$, that is the squared difference between the two summary statistics $\Phi_o$ and $\Phi_{\theta}$ in Example 4: $\Delta_{\theta} = (\Phi_o - \Phi_{\theta})^2$. Because of the use of simulated data, like the synthetic likelihood, the discrepancy $\Delta_{\theta}$ is a stochastic process. We visualize its distribution in Figure 4(a). The observed data $\mathbf{y}_o$ were the same as in Example 4.

For this simple example, we can compute the limiting approximate likelihood $\tilde{L}_u$ in Equation (25) in closed form,

$$\tilde{L}_u(\theta) \propto F\left(\sqrt{n}(\Phi_o - \theta) + \sqrt{nh}\right) - F\left(\sqrt{n}(\Phi_o - \theta) - \sqrt{nh}\right), \tag{27}$$

where $F(x)$ is the cumulative distribution function (cdf) of a standard normal random variable,

$$F(x) = \int_{-\infty}^{x} \frac{1}{\sqrt{2\pi}} \exp\left(-\frac{1}{2}u^2\right) \mathrm{d}u. \tag{28}$$

For small $nh$, $\tilde{L}_u(\theta)$ becomes proportional to the likelihood $L(\theta)$. This is visualized in Figure 4(b).[2] However, the probability to actually observe a realization of $\Delta_{\theta}$ which is below the threshold $h$ becomes vanishingly small. For realistic models, $\tilde{L}_u$ is not available in closed form but needs to be estimated. The vanishingly small probability indicates that the inference procedure will be computationally expensive when $\tilde{L}_u$ is estimated via the sample average approximation $\hat{L}_u^N$. ▲

**Example 7** (Approximate univariate likelihoods for the day care centers). In the model for bacterial infections in day care centers, the observed data were converted to summary statistics $\Phi_o$ by representing each day care center (binary matrix) with four statistics. This gives $4 \cdot 29 = 116$ summary statistics in total (see Numminen et al., 2013, for details).

---

2. Using $h = 0.1$ for illustrative purposes. For threshold choice in real applications, see Example 7 and Section 5.3.

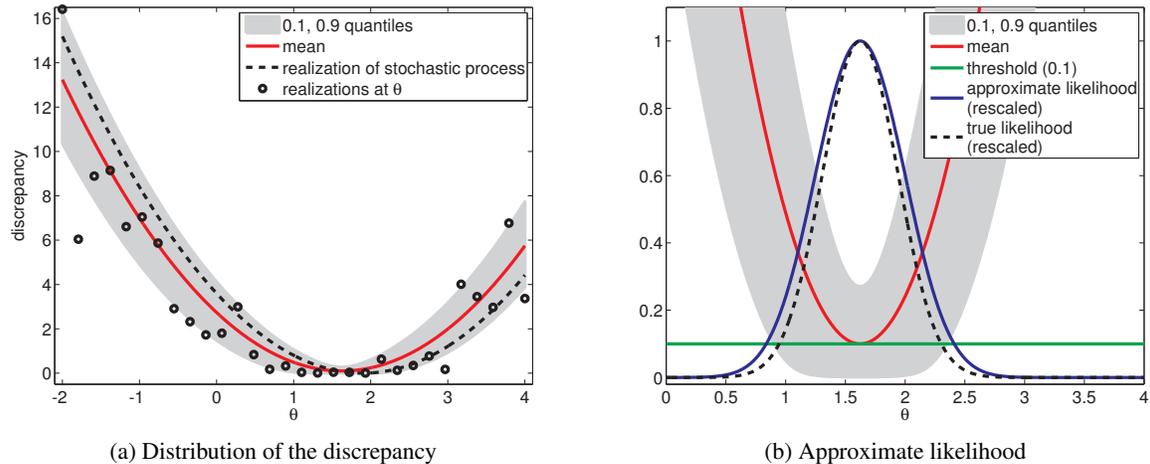

(a) Distribution of the discrepancy      (b) Approximate likelihood

Figure 4: Nonparametric approximation of the likelihood to estimate the mean $\theta$ of a Gaussian. The discrepancy $\Delta_\theta$ is the squared difference between the empirical means of the observed and simulated data. (a) The discrepancy is a random function. (b) The probability that the discrepancy is below some threshold $h$ approximates the likelihood. Note the different range of the axes.

Since the day care centers can be considered to be independent, the 29 observations can be used to estimate the distribution of the four statistics and their cdfs. Numminen et al. (2013) assessed the difference between $\Phi_\theta$ and $\Phi_o$ by the $L_1$ distance between the estimated cdfs. Each $L_1$ distance had its own uniform kernel and corresponding bandwidth, which means that a product kernel was used overall. We here work with a simplified discrepancy measure: The different scales of the four statistics were normalized by letting the maximal value of each of the four statistics be one for $\mathbf{y}_o$. The discrepancy $\Delta_\theta$ was then the $L_1$ norm between $\Phi_\theta$ and $\Phi_o$ divided by their dimension, $\Delta_\theta = 1/116||\Phi_\theta - \Phi_o||_1$.

Figure 5 shows the distributions of the discrepancies $\Delta_\theta$ if one of the three parameters is varied at a time. The results are for the real data used by Numminen et al. (2013). The parameters were varied on a grid around the (rounded) mean $(3.6, 0.6, 0.1)$ which was inferred by Numminen et al. (2013). The distributions were estimated using $N = 300$ realizations of $\Delta_\theta$ per parameter value. The red solid lines show the empirical average $\hat{J}^N$. The black lines with circles show $\hat{L}_u^N$ with bandwidths (thresholds) equal to the 0.1 quantile of the sampled discrepancies. While subjective, this is a customary choice (Marin et al., 2012). The thresholds were $h_\beta = 1.16$, $h_\Lambda = 1.18$, and $h_\theta = 1.20$, and are marked with green lines. It can be seen that the optima of $\hat{J}^N$ and $\hat{L}_u^N$ are attained at about the same parameter values which is advantageous because $\hat{J}^N$ is independent of kernel and bandwidth.

Since the functional form of $\hat{J}^N$ and its gradients are, however, not known, the minimization becomes a difficult problem in higher dimensions. We will show that the idea of

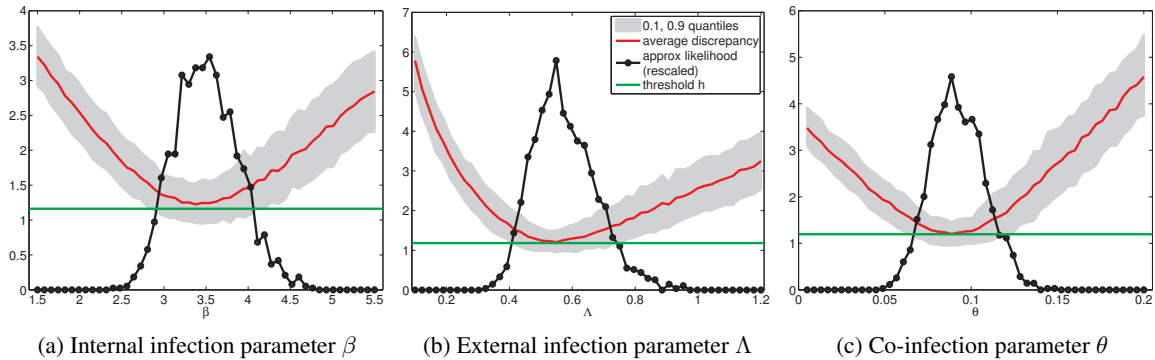

(a) Internal infection parameter $\beta$    (b) External infection parameter $\Lambda$    (c) Co-infection parameter $\theta$

Figure 5: Approximate likelihoods $\hat{L}_u^N$ and distributions of the discrepancy $\Delta_{\boldsymbol{\theta}}$ for the day care center example. The green horizontal lines indicate the thresholds used. The optima of the average discrepancies and the approximate likelihoods occur at about the same parameter values.

combining probabilistic modeling with optimization, which we mentioned in Example 5 for the log synthetic likelihood, is also helpful here. ▲

### 3.4 Relation between Nonparametric and Parametric Approximation

Kernel density estimation with Gaussian kernels is interesting for two reasons in the context of likelihood-free inference. First, the Gaussian kernel is positive definite, so that the estimated density is a member of a reproducing kernel Hilbert space. This means that more robust approximations of $p_{\Phi|\boldsymbol{\theta}}$ than the one in Equation (18) would exist (Kim and Scott, 2012), and that there might be connections to the inference approach of Fukumizu et al. (2013). Second, it allows us to embed the synthetic likelihood approach of Section 3.2 into the nonparametric approach of Section 3.3.

For the Gaussian kernel, we have that $K(\Phi_o, \Phi_{\boldsymbol{\theta}}) = K_g(\Phi_o - \Phi_{\boldsymbol{\theta}})$,

$$K_g(\Phi_o - \Phi_{\boldsymbol{\theta}}) = \frac{1}{(2\pi)^{p/2}} \frac{1}{|\det \mathbf{C}_{\boldsymbol{\theta}}|^{1/2}} \exp\left(-\frac{(\Phi_o - \Phi_{\boldsymbol{\theta}})^\top \mathbf{C}_{\boldsymbol{\theta}}^{-1}(\Phi_o - \Phi_{\boldsymbol{\theta}})}{2}\right), \tag{29}$$

where $\mathbf{C}_{\boldsymbol{\theta}}$ is a positive definite bandwidth matrix possibly depending on $\boldsymbol{\theta}$. The kernel $K_g$ corresponds to $\kappa = \kappa_g$ and $\Delta_{\boldsymbol{\theta}} = \Delta_{\boldsymbol{\theta}}^g$,

$$\kappa_g(u) = \frac{1}{(2\pi)^{p/2}} \exp\left(-\frac{u}{2}\right), \qquad \Delta_{\boldsymbol{\theta}}^g = \log|\det \mathbf{C}_{\boldsymbol{\theta}}| + (\Phi_o - \Phi_{\boldsymbol{\theta}})^\top \mathbf{C}_{\boldsymbol{\theta}}^{-1}(\Phi_o - \Phi_{\boldsymbol{\theta}}). \tag{30}$$

The function $\kappa_g$ is convex so that Equation (23) yields a lower bound for $\hat{L}^N(\boldsymbol{\theta}) = \hat{L}_g^N(\boldsymbol{\theta})$ and its logarithm,

$$\log \hat{L}_g^N(\boldsymbol{\theta}) \geq -\frac{p}{2}\log(2\pi) - \frac{1}{2}\hat{J}_g^N(\boldsymbol{\theta}), \tag{31}$$

$$\hat{J}_g^N(\boldsymbol{\theta}) = \mathrm{E}^N\left[\log|\det \mathbf{C}_{\boldsymbol{\theta}}| + (\Phi_o - \Phi_{\boldsymbol{\theta}})^\top \mathbf{C}_{\boldsymbol{\theta}}^{-1}(\Phi_o - \Phi_{\boldsymbol{\theta}})\right]. \tag{32}$$

We used the subscript "g" to highlight that $\hat{J}^N$ in Equation (23) is computed for the particular discrepancy $\Delta_{\boldsymbol{\theta}}^g$. The form of $\hat{J}_g^N$ is reminiscent of the log synthetic likelihood $\hat{\ell}_s^N$ in Equation (15). The following proposition shows that there is indeed a connection.

**Proposition 1** (Synthetic likelihood as lower bound). *For $\mathbf{C}_{\boldsymbol{\theta}} = \hat{\boldsymbol{\Sigma}}_{\boldsymbol{\theta}}$,*

$$\hat{\ell}_s^N(\boldsymbol{\theta}) = \frac{p}{2} - \frac{p}{2}\log(2\pi) - \frac{1}{2}\hat{J}_g^N(\boldsymbol{\theta}), \tag{33}$$

$$\log \hat{L}_g^N(\boldsymbol{\theta}) \geq -\frac{p}{2} + \hat{\ell}_s^N(\boldsymbol{\theta}) \tag{34}$$

The proposition is proved in Appendix A. It shows that maximizing the synthetic log-likelihood corresponds to maximizing a lower bound of a nonparametric approximation of the log likelihood. The proposition embeds the parametric approach to likelihood approximation conceptually in the nonparametric one and shows furthermore that $\hat{\ell}_s^N$ can be computed via an empirical expectation over $\Delta_{\boldsymbol{\theta}}^g$.

### 3.5 Posterior Inference using Sample Average Approximations of the Likelihood

Several computable approximations $\hat{L}$ of the likelihood $L$ were constructed in the previous two sections. Table 1 provides an overview. Intractable expectations were replaced with sample averages using $N$ simulated data sets which we denoted by the superscript "$N$" in the symbols for the approximations.

Wood (2010) used the synthetic likelihood $\hat{L}_s^N$ together with a Metropolis MCMC algorithm for posterior computations. We here focus on posterior inference via importance sampling. Using $\hat{L}_u^N$ as $\hat{L}$ in Equation (11), we have

$$\mathrm{E}(g(\boldsymbol{\theta})|\Phi_o) \approx \sum_{m=1}^M g(\boldsymbol{\theta}^{(m)})\hat{w}_u^{(m)}, \qquad \hat{w}_u^{(m)} = \frac{\sum_{j=1}^N \chi_{[0,h)}(\Delta_{\boldsymbol{\theta}}^{(jm)})\frac{p_{\boldsymbol{\theta}}(\boldsymbol{\theta}^{(m)})}{q(\boldsymbol{\theta}^{(m)})}}{\sum_{i=1}^M \sum_{j=1}^N \chi_{[0,h)}(\Delta_{\boldsymbol{\theta}}^{(ji)})\frac{p_{\boldsymbol{\theta}}(\boldsymbol{\theta}^{(i)})}{q(\boldsymbol{\theta}^{(i)})}}, \tag{35}$$

where $\chi_{[0,h)}$ is the indicator function of the interval $[0,h)$ as before, and the $\Delta_{\boldsymbol{\theta}}^{(jm)}$, $j = 1,\ldots,N$, are the observed discrepancies for the sampled parameter $\boldsymbol{\theta}^{(m)} \sim q(\boldsymbol{\theta})$. Instead of sampling several discrepancies for the same $\boldsymbol{\theta}^{(m)}$, sampling $M'$ pairs $(\Delta_{\boldsymbol{\theta}}^{(i)}, \boldsymbol{\theta}^{(i)})$ with $N = 1$ is also possible and corresponds to an asymptotically equivalent solution. With $q = p_{\boldsymbol{\theta}}$, the approximation is a Nadaraya–Watson kernel estimate of the conditional expectation (see, for example, Wasserman, 2004, Chapter 21).

Approximate Bayesian computation (ABC) is intrinsically linked to kernel density estimation and kernel regression (Blum, 2010). A basic ABC rejection sampler (Pritchard et al., 1999; Marin et al., 2012, Algorithm 2) is obtained from Equation (35) with $N = 1$, $q = p_{\boldsymbol{\theta}}$, and $\Delta_{\boldsymbol{\theta}} = ||\Phi_o - \Phi_{\boldsymbol{\theta}}||$ where $||.||$ is some norm. Approximate samples from the posterior pdf of $\boldsymbol{\theta}$ given $\Phi_o$ can thus be obtained by retaining those $\boldsymbol{\theta}^{(m)}$ for which the $\Phi_{\boldsymbol{\theta}}^{(m)}$ are within distance $h$ from $\Phi_o$. In an iterative approach, the accepted particles can be used to define the auxiliary pdf $q(\boldsymbol{\theta})$ of the next iteration by letting it be a mixture of Gaussians with weights $\hat{w}_u^{(m)}$, center points $\boldsymbol{\theta}^{(m)}$, and a covariance determined by the $\boldsymbol{\theta}^{(m)}$ (Beaumont et al., 2009). This gives the population Monte Carlo (PMC) ABC algorithm (Marin et al., 2012, Algorithm 4). Related sequential Monte Carlo (SMC) ABC algorithms were proposed by Sisson et al. (2007) and Toni et al. (2009). Working with $q = p_{\boldsymbol{\theta}}$, Beaumont et al. (2002) introduced ABC with more general kernels, which corresponds to using $\hat{L}_\kappa^N$ instead of $\hat{L}_u^N$.

Example 6 showed that approximating the likelihood via sample averages is computationally expensive because of the required small thresholds. The auxiliary pdf $q(\boldsymbol{\theta})$ specifies where in the parameter space the likelihood is predominantly evaluated. The following example shows that avoiding regions in the parameter space where the likelihood is vanishingly small allows for considerable computational savings.

**Example 8** (Univariate approximate posteriors for the day care centers). For the inference of the model of bacterial infections in day care centers, Numminen et al. (2013) used uniform priors for the parameters $\beta \in (0, 11)$, $\Lambda \in (0, 2)$, and $\theta \in (0, 1)$. The likelihoods $\hat{L}_u^N$ shown in Figure 5 are thus proportional to the posterior pdfs. The posterior pdfs of the univariate unknowns are conditional on the remaining fixed parameters. For example, the posterior pdf for $\beta$ is conditional on $(\Lambda, \theta) = (\Lambda_o, \theta_o) = (0.6, 0.1)$. In Section 7, we consider inference of all three parameters at the same time.

In Figure 5, each parameter is evaluated on a sub-interval of the domain of the prior. The sub-intervals were chosen such that the far tails of the likelihoods were excluded. Parameter $\beta$, for example, was evaluated on the interval $(1.5, 5.5)$ only. Evaluating the discrepancy $\Delta_{\boldsymbol{\theta}}$ on the complete interval $(0, 11)$ is not very meaningful since the probability that it is above the chosen threshold is vanishingly small outside the interval $(1.5, 5.5)$. In fact, out of $M = 5{,}000$ discrepancies $\Delta_{\boldsymbol{\theta}}$ which we simulated for $\beta$ uniformly on $(0, 11)$, not a single one was accepted for $\beta \notin (1.5, 5.5)$. Hence, taking for instance a uniform distribution on $(1.5, 5.5)$ instead of the prior as auxiliary distribution leads to considerable computational savings. Motivated by this, we propose a method which automatically avoids regions in the parameter space where the likelihood is vanishingly small. ▲

## 4. Computational Difficulties in the Standard Inference Approach

We have seen that the approximate likelihood functions $\hat{L}(\boldsymbol{\theta})$ which are used to infer simulator-based statistical models are stochastic processes indexed by the model parameters $\boldsymbol{\theta}$. Their properties, in particular their functional form and gradients, are generally not known; they behave like stochastic black-box functions. The stochasticity is due to the use of simulations to approximate intractable expectations. In the standard approach presented in the previous section, the expectations are approximated by sample averages so that a single evaluation of $\hat{L}$ requires the simulation of $N$ data sets. The standard approach makes minimal assumptions but suffers from a couple of limiting factors.

1. There is an inherent trade-off between computational and statistical efficiency: Reducing $N$ reduces the computational cost of the inference methods, but it can also decrease the accuracy of the estimates (Figure 2).

2. For finite $N$, the approximate likelihoods may not be smooth (Figure 3).

3. Simulating $N$ data sets uniformly in the parameter space is an inefficient use of computational resources and particularly costly if simulating a single data set already takes a long time. In some regions in the parameter space, far fewer simulations suffice to conclude that it is very unlikely for the approximate likelihood to take a significant value (Figures 2 to 5).

## 5. Framework to Increase the Computationally Efficiency

We present a framework which combines optimization with probabilistic modeling in order to increase the efficiency of likelihood-free inference of simulator-based statistical models.

### 5.1 From Sample Average to Regression Based Approximations

The standard approach to obtain a computable approximate likelihood function $\hat{L}$ relies on sample averages, yielding the parametric approximation $\hat{L}_s^N = \exp(\hat{\ell}_s^N)$ in Equation (15) or the nonparametric approximation $\hat{L}_\kappa^N$ in Equation (21). The approximations are computable versions of $\tilde{L}_s = \exp(\tilde{\ell}_s)$ in Equation (13) and $\tilde{L}_\kappa$ in Equation (22), which both involve intractable expectations. But sample averages are not the only way to approximate the intractable expectations. We here consider approximations based on regression.

Equation (22) shows that $\tilde{L}_\kappa(\boldsymbol{\theta})$ has a natural interpretation as a regression function where the model parameters $\boldsymbol{\theta}$ are the covariates (the independent variables) and $\kappa(\Delta_{\boldsymbol{\theta}})$ is the response variable. The expectation can thus also be approximated by solving a regression problem. Further, $\hat{J}^N$ in Equation (23) can be seen as the sample average approximation of the regression function $J(\boldsymbol{\theta})$,

$$J(\boldsymbol{\theta}) = \mathrm{E}\left[\Delta_{\boldsymbol{\theta}}\right], \tag{36}$$

where the discrepancy $\Delta_{\boldsymbol{\theta}}$ is the response variable. The arguments which we used to show that $\hat{J}^N$ provides a lower bound for $\hat{L}_\kappa^N$ carry directly over to $J$ and $\tilde{L}_\kappa$: $J$ provides a lower bound for $\tilde{L}_\kappa$ if $\kappa$ is convex or the uniform kernel.

Proposition 1 establishes a relation between the sample average quantities $\hat{J}_g^N$ in Equation (32) and $\hat{\ell}_s^N$ in Equation (15). In the proof of the proposition in Appendix A, we show that the relation extends to the limiting quantities $J_g(\boldsymbol{\theta}) = \mathrm{E}\left[\Delta_{\boldsymbol{\theta}}^g\right]$ and $\tilde{\ell}_s$ in Equation (13). Thus, for $\mathbf{C}_{\boldsymbol{\theta}} = \boldsymbol{\Sigma}_{\boldsymbol{\theta}}$ and up to constants and the sign, $\tilde{\ell}_s(\boldsymbol{\theta})$ can be seen as a regression function with the particular discrepancy $\Delta_{\boldsymbol{\theta}}^g$ as the response variable.

We next discuss the general strategy to infer the regression functions while avoiding unnecessary computations. For nonparametric approximations to the likelihood, inferring $J$ is simpler than inferring $\tilde{L}_\kappa$ since the function $\kappa$ and its corresponding bandwidth do not need to be chosen. We thus propose to first infer the regression function $J$ of the discrepancies and then, in a second step, to leverage the obtained solution to infer $\tilde{L}_\kappa$. For the parametric approximation to the likelihood, this extra step is not needed since $J_g$ is a special instance of the regression function $J$.

### 5.2 Inferring the Regression Function of the Discrepancies

Inferring $J(\boldsymbol{\theta})$ via regression requires training data in the form of tuples $(\boldsymbol{\theta}^{(i)}, \Delta_{\boldsymbol{\theta}}^{(i)})$. Since we are mostly interested in the region of the parameter space where $\Delta_{\boldsymbol{\theta}}$ tends to be small, we propose to actively construct the training data such that they are more densely clustered around the minimizer of $J(\boldsymbol{\theta})$. As $J(\boldsymbol{\theta})$ is unknown in the first place, our proposal amounts to performing regression and optimization at the same time: Given an initial guess that the minimizer is in some bounded subset of the parameter space, we can sample some evidence $\mathcal{E}^{(t)}$ of the relation between $\boldsymbol{\theta}$ and $\Delta_{\boldsymbol{\theta}}$,

$$\mathcal{E}^{(t)} = \left\{(\boldsymbol{\theta}^{(1)}, \Delta_{\boldsymbol{\theta}}^{(1)}), \dots, (\boldsymbol{\theta}^{(t)}, \Delta_{\boldsymbol{\theta}}^{(t)})\right\}, \tag{37}$$

and use this evidence to obtain an estimate $\hat{J}^{(t)}$ of $J$ via regression. The estimated $\hat{J}^{(t)}$ and some measurement of uncertainty about it can then be used to produce a new guess about the potential location of the minimizer, from where the process re-starts. In some cases, it may be advantageous to include the prior pdf of the parameters in the process. We explore this topic in Appendix B.

The evidence set $\mathcal{E}^{(t)}$ grows at every iteration and we may stop at $t = T$. The value of $T$ can be chosen based on computational considerations, by checking whether the learned model predicts the acquired points reasonably well, or by monitoring the change in the minimizer $\hat{\boldsymbol{\theta}}_J^{(t)}$ of $\hat{J}^{(t)}$ as the evidence set grows,

$$\hat{\boldsymbol{\theta}}_J^{(t)} = \operatorname{argmin}_{\boldsymbol{\theta}} \hat{J}^{(t)}(\boldsymbol{\theta}). \tag{38}$$

Given our examples so far, it is further reasonable to assume that $J$ is a smooth function. Even for the Ricker model, the mean objective was smooth although the individual realizations were not (Figure 3). The smoothness assumption about $J$ can be used in the regression and enables its efficient minimization.

For the special case where $\tilde{\ell}_s$ is the target, several observed values of $\Delta_{\boldsymbol{\theta}} = \Delta_{\boldsymbol{\theta}}^g$ may be available for any given $\boldsymbol{\theta}^{(i)}$. This is because the covariance matrix $\boldsymbol{\Sigma}_{\boldsymbol{\theta}}$ may be still estimated as a sample average so that multiple simulated summary statistics, and hence discrepancies, are available per $\boldsymbol{\theta}^{(i)}$. They can be used as discussed above with the only minor modification that the training data are updated with several tuples at a time. But it is also possible to only use the average value of the observed discrepancies, which amounts to using the observed values of $\hat{\ell}_s^N$ for training. The estimated regression function $\hat{J}^{(t)}$ provides an estimate for $\tilde{\ell}_s$ in either case. We denote the estimate by $\hat{\ell}_s^{(t)}$ and the corresponding estimate of $\tilde{L}_s$ by $\hat{L}_s^{(t)}$.

Combining nonlinear regression with the acquisition of new evidence in order to optimize a black-box function is known as Bayesian optimization (see, for example, Brochu et al., 2010). We can thus leverage results from Bayesian optimization to implement the proposed approach, which we will do in Section 6.

## 5.3 Inferring the Regression Function for Nonparametric Likelihood Approximation

The evidence set $\mathcal{E}^{(t)}$ can be used in two possible ways in the nonparametric setting: The first possibility is to compute for each $\Delta_{\boldsymbol{\theta}}^{(i)}$ in $\mathcal{E}^{(t)}$ the value $\kappa^{(i)} = \kappa(\Delta_{\boldsymbol{\theta}}^{(i)})$ and to thereby produce a new evidence set which can be used to approximate $\tilde{L}_\kappa$ by fitting a regression function. The second possibility is to estimate a probabilistic model of $\Delta_{\boldsymbol{\theta}}$ from the evidence $\mathcal{E}^{(t)}$. The estimated model can be used to approximate $\tilde{L}_\kappa$ by replacing the expectation in Equation (22) with the expectation under the model. We denote either approximation by $\hat{L}_\kappa^{(t)}$ where the superscript "$(t)$" indicates that the approximation was obtained via regression with $t$ training points. Since $\mathcal{E}^{(t)}$ is such that the approximation of the regression function is accurate where it takes small values, the approximation of $\tilde{L}_\kappa$ will be accurate where it takes large values, that is, in the modal areas.

For nonparametric likelihood approximation, kernels and bandwidths need to be selected (see Section 3.3). The choice of the kernel is generally thought to be less critical than the choice of the bandwidth (Wand and Jones, 1995). Bandwidth selection has received considerable attention in the literature on kernel density estimation (for an introduction, see, for example, Wand and Jones, 1995). The results from that literature are, however, not straightforwardly applicable to our work: We may only be given a certain discrepancy measure $\Delta_{\boldsymbol{\theta}}$ without underlying summary statistics $\Phi_{\boldsymbol{\theta}}$ (Gutmann et al., 2014). And even if the discrepancy $\Delta_{\boldsymbol{\theta}}$ is constructed via summary statistics, the kernel density estimate is only evaluated at $\Phi_o$ which is kept fixed while $\boldsymbol{\theta}$ is varied. Furthermore, we usually only have very few observations available for any given $\boldsymbol{\theta}$ which is generally not the case in kernel density estimation. These differences warrant further investigations into which extent the bandwidth selection methods from the kernel density estimation literature are applicable to likelihood-free inference. We focus in this paper on the uniform kernel and generally choose $h$ via the quantiles of the $\Delta_{\boldsymbol{\theta}}^{(i)}$, which is common practice in approximate Bayesian computation (see, for example, Marin et al., 2012). The approximate likelihood function for the uniform kernel will be denoted by $\hat{L}_u^{(t)}$.

### 5.4 Benefits and Limitations of the Proposed Approach

The difference between the proposed approach and the standard approach to likelihood-free inference of simulator-based statistical models lies in the way the intractable $J$ and $\tilde{L}$ are approximated. We use regression with actively acquired training data while the standard approach relies on computing sample averages. Our approach allows to incorporate a smoothness assumption about $J$ and $\tilde{L}$ in the region of their optima. The smoothness assumption allows to "share" observed $\Delta_{\boldsymbol{\theta}}$ among multiple $\boldsymbol{\theta}$ which suggests that fewer $\Delta_{\boldsymbol{\theta}}^{(i)}$, that is, fewer simulated data sets $\mathbf{y}_{\boldsymbol{\theta}}^{(i)}$, are needed to reach a certain level of accuracy. A second benefit of the proposed approach is that it directly targets the region in the parameter space where the discrepancy $\Delta_{\boldsymbol{\theta}}$ tends to be small, which is very important if simulating data sets is time consuming.

Regression and deciding on the training data are not free of computational cost. While the additional expense is often justified by the net savings made, it goes without saying that if simulating the model is very cheap, methods for regression and decision making need to be used which are not disproportionately costly. Furthermore, prioritizing the low-discrepancy areas of the parameter space is often meaningful, but it also implies that the tails of the likelihood (posterior) will not be as well approximated as the modal areas. The proposed approach thus had to be modified if the computation of small probability events was of primary interest.

Section 4 lists three computational difficulties occurring in the standard approach. Our approach addresses the smoothness issues via smooth regression. The inefficient use of resources is addressed by focusing on regions in the parameter space where $\Delta_{\boldsymbol{\theta}}$ tends to be small. The trade-off between computational and statistical performance is still present but in modified form: The trade-off is the size of the training set $\mathcal{E}^{(t)}$ used in the regression. The regression functions can be estimated more accurately as the size of the training set grows but this also requires more computation. The size of the training set as trade-off parameter has the advantage that we are free to choose in which areas of the parameter space we would like to approximate the regression function more accurately and in which areas an accurate approximation is not needed. This is in contrast to the standard approach where a computational cost of $N$ simulated data sets needs to be paid per $\boldsymbol{\theta}$ irrespective of its value.

## 6. Implementing the Framework with Bayesian Optimization

We start with introducing Bayesian optimization and then use it to implement our framework. This is followed by a discussion of possible extensions.

### 6.1 Brief Introduction to Bayesian Optimization

We briefly introduce the elements of Bayesian optimization which are needed in the paper. A more thorough introduction can be found in the review articles by Jones (2001) and Brochu et al. (2010). While the presented version of Bayesian optimization is rather straightforward and textbook-like, our framework can also be implemented with more advanced versions, see Section 6.4.

Bayesian optimization comprises a set of methods to minimize black-box functions $f(\boldsymbol{\theta})$. With a black-box function, we mean a function which we can evaluate but whose form and gradients are unknown. The basic idea in Bayesian optimization is to use a probabilistic model of $f$ to select points where the objective is evaluated, and to use the obtained values to update the model by Bayes' theorem.

The objective $f$ is often modeled as a Gaussian process which is also done in this paper: We assume that $f$ is a Gaussian process with prior mean function $m(\boldsymbol{\theta})$ and covariance function $k(\boldsymbol{\theta}, \boldsymbol{\theta}')$ subject to additive Gaussian observation noise with variance $\sigma_n^2$. The joint distribution of $f$ at any $t$ points $\boldsymbol{\theta}^{(1)}, \ldots, \boldsymbol{\theta}^{(t)}$ is thus assumed Gaussian with mean $\mathbf{m}_t$ and covariance $\mathbf{K}_t$,

$$(f^{(1)}, \ldots, f^{(t)})^\top \sim \mathcal{N}(\mathbf{m}_t, \mathbf{K}_t), \tag{39}$$

$$\mathbf{m}_t = \begin{pmatrix} m(\boldsymbol{\theta}^{(1)}) \\ \vdots \\ m(\boldsymbol{\theta}^{(t)}) \end{pmatrix}, \qquad \mathbf{K}_t = \begin{pmatrix} k(\boldsymbol{\theta}^{(1)}, \boldsymbol{\theta}^{(1)}) & \ldots & k(\boldsymbol{\theta}^{(1)}, \boldsymbol{\theta}^{(t)}) \\ \vdots & & \vdots \\ k(\boldsymbol{\theta}^{(t)}, \boldsymbol{\theta}^{(1)}) & \ldots & k(\boldsymbol{\theta}^{(t)}, \boldsymbol{\theta}^{(t)}) \end{pmatrix} + \sigma_n^2 \mathbf{I}_t. \tag{40}$$

We used $f^{(i)}$ to denote $f(\boldsymbol{\theta}^{(i)})$ and $\mathbf{I}_t$ is the $t \times t$ identity matrix. While other choices are possible, we assume that $m(\boldsymbol{\theta})$ is either a constant or a sum of convex quadratic polynomials in the elements $\theta_j$ of $\boldsymbol{\theta}$, cross-terms were not included, and that $k(\boldsymbol{\theta}, \boldsymbol{\theta}')$ is a squared exponential covariance function,

$$m(\boldsymbol{\theta}) = \sum_j a_j \theta_j^2 + b_j \theta_j + c, \qquad k(\boldsymbol{\theta}, \boldsymbol{\theta}') = \sigma_f^2 \exp\left(\sum_j \frac{1}{\lambda_j^2}(\theta_j - \theta_j')^2\right). \tag{41}$$

These are standard choices (see, for example, Rasmussen and Williams, 2006, Chapter 2). Since we are interested in minimization, we constrain the $a_j$ to be non-negative. In the last equation, $\theta_j$ and $\theta_j'$ are the elements of $\boldsymbol{\theta}$ and $\boldsymbol{\theta}'$, respectively, $\sigma_f^2$ is the signal variance, and the $\lambda_j$ are the characteristic length scales. The length scales control the amount of correlation between $f(\boldsymbol{\theta})$ and $f(\boldsymbol{\theta}')$, in other words, they control the wiggliness of the realizations of the Gaussian process. The signal variance is the marginal variance of $f$ at a point $\boldsymbol{\theta}$ if the observation noise was zero.

The quantities $a_j$, $b_j$, $c$, $\sigma_f^2$, $\lambda_j$, and $\sigma_n^2$ are hyperparameters. For the results in this paper, they were learned by maximizing the leave-one-out log predictive probability (a form of cross-validation, see Rasmussen and Williams, 2006, Section 5.4.2). The hyperparameters were slowly updated as new data were acquired, as done in previous work, for example by Wang et al. (2013). This yielded satisfactory results but there are several alternatives, including Bayesian methods to learn the hyperparameters (for an overview, see Rasmussen and Williams, 2006, Chapter 5), and we did not perform any systematic comparison.

Given evidence $\mathcal{E}_f^{(t)} = \{(\boldsymbol{\theta}^{(1)}, f^{(1)}), \ldots, (\boldsymbol{\theta}^{(t)}, f^{(t)})\}$, the posterior pdf of $f$ at a point $\boldsymbol{\theta}$ is Gaussian with posterior mean $\mu_t(\boldsymbol{\theta})$ and posterior variance $v_t(\boldsymbol{\theta}) + \sigma_n^2$,

$$f(\boldsymbol{\theta})|\mathcal{E}_f^{(t)} \sim \mathcal{N}(\mu_t(\boldsymbol{\theta}), v_t(\boldsymbol{\theta}) + \sigma_n^2), \tag{42}$$

where (see, for example, Rasmussen and Williams, 2006, Section 2.7),

$$\mu_t(\boldsymbol{\theta}) = m(\boldsymbol{\theta}) + \mathbf{k}_t(\boldsymbol{\theta})^\top \mathbf{K}_t^{-1}(\mathbf{f}_t - \mathbf{m}_t), \qquad v_t(\boldsymbol{\theta}) = k(\boldsymbol{\theta}, \boldsymbol{\theta}) - \mathbf{k}_t(\boldsymbol{\theta})^\top \mathbf{K}_t^{-1} \mathbf{k}_t(\boldsymbol{\theta}), \tag{43}$$

$$\mathbf{f}_t = (f^{(1)}, \ldots, f^{(t)})^\top, \qquad \mathbf{k}_t(\boldsymbol{\theta}) = (k(\boldsymbol{\theta}, \boldsymbol{\theta}^{(1)}), \ldots, k(\boldsymbol{\theta}, \boldsymbol{\theta}^{(t)}))^\top. \tag{44}$$

The posterior mean $\mu_t$ emulates $f$ and can be minimized with powerful gradient-based optimization methods.

The evidence set can be augmented by selecting a new point $\boldsymbol{\theta}^{(t+1)}$ where $f$ is next evaluated. The point is chosen based on the posterior distribution of $f$ given $\mathcal{E}_f^{(t)}$. While other choices are equally possible, we use the acquisition function $\mathcal{A}_t(\boldsymbol{\theta})$ to select the next point,

$$\mathcal{A}_t(\boldsymbol{\theta}) = \mu_t(\boldsymbol{\theta}) - \sqrt{\eta_t^2 v_t(\boldsymbol{\theta})}, \tag{45}$$

where $\eta_t^2 = 2\log[t^{d/2+2}\pi^2/(3\epsilon_\eta)]$ with $\epsilon_\eta$ being a small constant (we used $\epsilon_\eta = 0.1$). This acquisition function is known as the lower confidence bound selection criterion (Cox and John, 1992, 1997; Srinivas et al., 2010, 2012).[3] Classically, $\boldsymbol{\theta}^{(t+1)}$ is chosen deterministically as the minimizer of $\mathcal{A}_t(\boldsymbol{\theta})$. The minimization of $\mathcal{A}_t(\boldsymbol{\theta})$ yields a compromise between exploration and exploitation: Minimization of the posterior mean $\mu_t(\boldsymbol{\theta})$ corresponds to exploitation of the current belief and ignores its uncertainty. Minimization of $-\sqrt{v_t(\boldsymbol{\theta})}$, on the other hand, corresponds to exploration where we seek a point where we are uncertain about $f$. The coefficient $\eta_t$ implements the trade-off between these two desiderata.

There is usually no restriction that $\boldsymbol{\theta}^{(t+1)}$ must be different from previously acquired $\boldsymbol{\theta}^{(t)}$. We found, however, that this may result in a poor exploration of the parameter space (see Figure 7 and Example 10 below). Employing a stochastic acquisition rule avoids getting stuck at one point. We used the simple heuristic that $\boldsymbol{\theta}^{(t+1)}$ is sampled from a Gaussian with diagonal covariance matrix and mean equal to the minimizer of the acquisition function. The standard deviations were determined by finding the end-points of the interval where the acquisition function was within a certain (relative) tolerance. Other stochastic acquisition rules, like for example Thompson sampling (Thompson, 1933; Chapelle and Li, 2011; Russo and Van Roy, 2014), could alternatively be used.

The algorithm was initialized with an evidence set $\mathcal{E}_f^{(t_0)}$ where the parameters $\boldsymbol{\theta}^{(1)}, \ldots, \boldsymbol{\theta}^{(t_0)}$ were chosen as a Sobol quasi-random sequence (see, for example, Niederreiter, 1988). Compared to uniformly distributed (pseudo) random numbers, the Sobol sequence covers the parameter space in a more even fashion. This kind of initialization is, however, not critical to our approach, and only few initial points were used in our simulations.

---

3. In the literature, maximization instead of minimization problems are often considered. For maximization problems, the acquisition function becomes $\mu_t(\boldsymbol{\theta}) + \sqrt{\eta_t^2 v_t(\boldsymbol{\theta})}$ and needs to be maximized. The formula for $\eta_t^2$ is used in the review by Brochu et al. (2010) and is part of Theorem 2 of Srinivas et al. (2010).

### 6.2 Inferring the Regression Function of the Discrepancies

Letting $f(\boldsymbol{\theta}) = \Delta_{\boldsymbol{\theta}}$, Bayesian optimization yields immediately an estimate of $J(\boldsymbol{\theta})$ in Equation (36). Since $\Delta_{\boldsymbol{\theta}}$ is non-negative, working with $f = \log \Delta_{\boldsymbol{\theta}}$ seems to be theoretically more sound. In practice, however, both approaches were found to work well, albeit we do not aim at any systematic comparison here. If $f = \Delta_{\boldsymbol{\theta}}$, the estimate $\hat{J}^{(t)}$ of $J$ is given by the posterior mean $\mu_t$, and if $f = \log \Delta_{\boldsymbol{\theta}}$, the estimate is given by the mean of a log-normal random variable,

$$\hat{J}^{(t)}(\boldsymbol{\theta}) = \begin{cases} \mu_t(\boldsymbol{\theta}) & \text{if } f(\boldsymbol{\theta}) = \Delta_{\boldsymbol{\theta}}, \\ \exp\left(\mu_t(\boldsymbol{\theta}) + \frac{1}{2}(v_t(\boldsymbol{\theta}) + \sigma_n^2)\right) & \text{if } f(\boldsymbol{\theta}) = \log \Delta_{\boldsymbol{\theta}}. \end{cases} \tag{46}$$

As discussed in Section 5.2, in the parametric approach to likelihood approximation, $\hat{J}^{(t)}$ equals the computable approximation $\hat{\ell}_s^{(t)}$ of $\bar{\ell}_s$.

We illustrate the basic principles of Bayesian optimization in Example 9 below. In Example 10, we illustrate log-Gaussian modeling and the stochastic acquisition rule.

**Example 9** (Bayesian optimization to infer the mean of a Gaussian). For inference of the mean of a univariate Gaussian, the squared difference of the empirical means was used as the discrepancy measure $\Delta_\theta$, as in Example 6. We modeled the discrepancy $\Delta_\theta$ as a Gaussian process with constant prior mean and performed Bayesian optimization with the deterministic acquisition rule. Figure 6 shows the first iterations: When only a single observation of $\Delta_\theta$ is available, $t = 1$ and $\Delta_\theta$ is believed to be constant but there is considerable uncertainty about it (upper-left panel). The posterior distribution of the Gaussian process yields the acquisition function $\mathcal{A}_1(\boldsymbol{\theta})$ according to Equation (45) (curve in magenta). Its minimization gives the value $\theta^{(2)}$ where $\Delta_\theta$ is evaluated next (blue rectangle). After including the observed value of $\Delta_\theta$ into the evidence set, $t = 2$ and the posterior distribution of the Gaussian process is re-calculated using Equation (42), that is, the belief about $\Delta_\theta$ is updated using Bayes' theorem (upper-right panel). The updated belief becomes the current belief and the process restarts. A movie showing the process over several iterations is available at `http://www.cs.helsinki.fi/u/gutmann/material/BOLFI/movies/Gauss.avi`. ▲

**Example 10** (Bayesian optimization to infer the growth rate in the Ricker model). Example 5 introduced the synthetic likelihood for the Ricker model. We have seen that individual realizations of $\hat{\ell}_s^N$ are rather noisy, in particular for $N = 50$, but that their average, which represents an estimate of $\tilde{\ell}_s$, is smooth with its optimum in the right region (Figure 3). We here obtain estimates $\hat{\ell}_s^{(t)}$ of $\tilde{\ell}_s$ with Bayesian optimization. The maximal training data are $T = 150$ tuples $(\log r, \hat{\ell}_s^N)$, where the first nine are from the initialization. The log synthetic likelihood was computed using code of Wood (2010) which only returned $\hat{\ell}_s^N$ and not the multiple discrepancies prior to averaging.

Figures 7(a) and (b) show $-\hat{\ell}_s^{(t)}$ after initialization without and with log-transformation, respectively (black solid lines). In both cases, we used a quadratic prior mean function. The estimated limiting negative log synthetic likelihood $-\bar{\ell}_s$ from Figure 3 is shown in red for

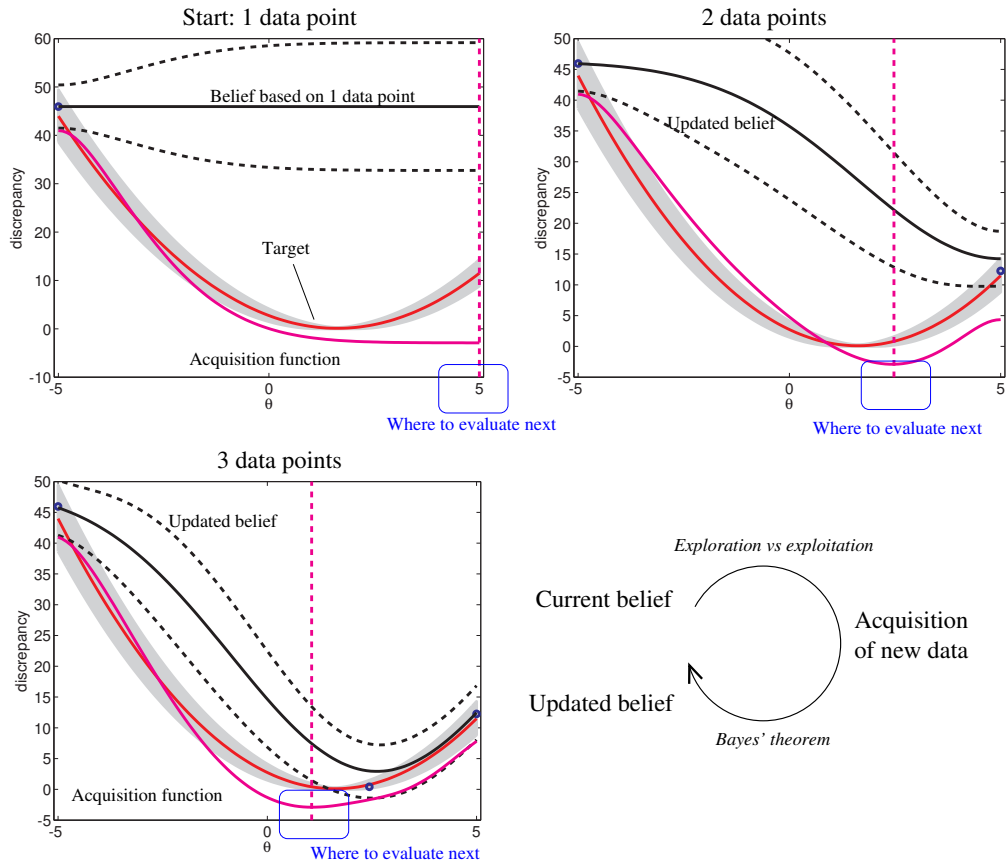

Figure 6: The first iterations of Bayesian optimization to estimate the mean of a Gaussian. The distribution of $\Delta_\theta$ and its regression function $J(\theta)$ are reproduced from Figure 4 for reference (labeled "Target"). Bayesian optimization consists in acquiring new data based on the current belief, followed by an update of the belief by Bayes' theorem. The acquisition of new data is based on an acquisition function which implements a trade-off between exploration and exploitation. Exploitation after two data points would consist in evaluating the objective again at $\theta = 5$. Exploration would consist in evaluating it where the posterior variance is large, that is, somewhere between minus five and zero. The point selected (blue rectangle) strikes a compromise between the two extremes.

reference. Figure 7(c) shows that the deterministic decision rule can lead to acquisitions with very little spatial exploration. The reason for the poor exploration is presumably the rather large variance of $\hat{\ell}_s^N$ for $N = 50$. Working with a log-Gaussian process leads to a better exploration and also to a better approximation (Figure 7(d)). The acquisitions happen, however, still in a cluster-like manner, which can also be seen in Figure 14 in Appendix C where we provide a more detailed analysis. Working with a stochastic decision rule leads to acquired points which are spread out more evenly in the area of interest. This results in both more stable and more accurate approximations (Figures 7(e–f) and Figure 15 in Appendix C). ▲

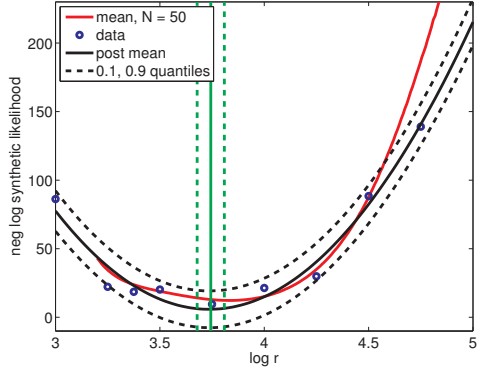

(a) Gaussian process model, initial

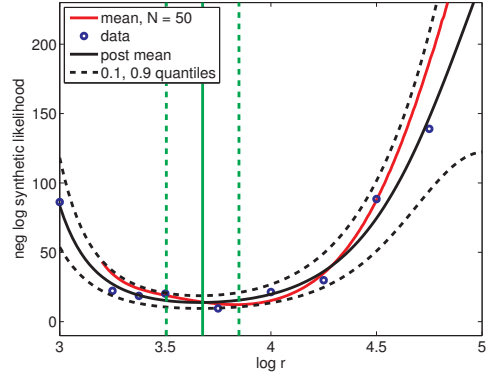

(b) log-Gaussian process model, initial

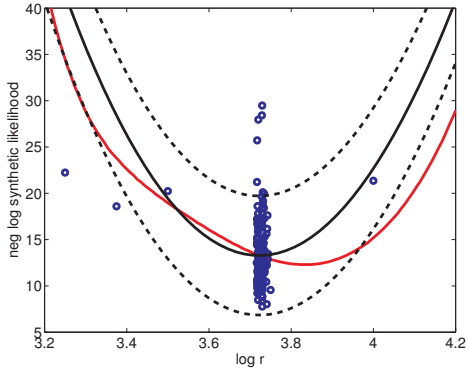

(c) GP, deterministic decision

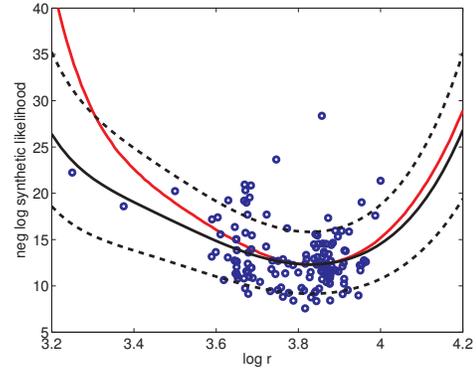

(d) log GP, deterministic decision

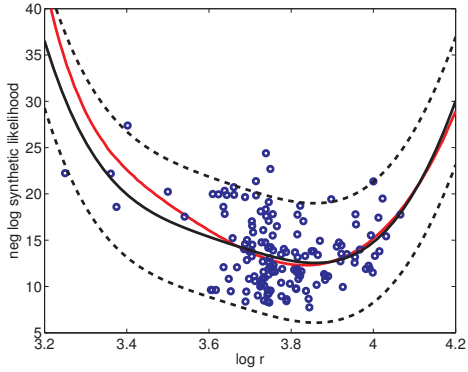

(e) GP, stochastic decision

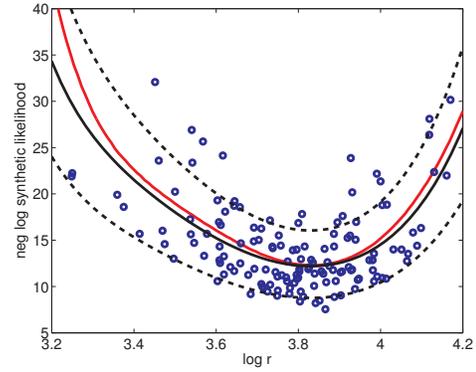

(f) log GP, stochastic decision

Figure 7: Approximation of the limiting negative log synthetic likelihood $-\tilde{\ell}_s$ for the Ricker example. The approximations are shown as black solid curves. The black dashed curves indicate the variability of $-\hat{\ell}_s^N$, and the red curves show $-\tilde{\ell}_s$ from Figure 3 for reference. (a–b) The approximation after initialization with 9 data points. The green vertical lines indicate the minimizer of the acquisition function. The dashed vertical lines show the mean plus-or-minus one standard deviation in the stochastic decision rule. (c–f) The approximations are based on 150 data points (blue circles).

### 6.3 Model-Based Nonparametric Likelihood Approximation

Bayesian optimization yields a probabilistic model for the discrepancy $\Delta_{\boldsymbol{\theta}}$. As discussed in Section 5.3, we can use this model to obtain the computable likelihood approximation $\hat{L}_u^{(t)}$,

$$\hat{L}_u^{(t)}(\boldsymbol{\theta}) \propto \begin{cases} F\left(\frac{h - \mu_t(\boldsymbol{\theta})}{\sqrt{v_t(\boldsymbol{\theta}) + \sigma_n^2}}\right) & \text{if } f(\boldsymbol{\theta}) = \Delta_{\boldsymbol{\theta}}, \\ F\left(\frac{\log h - \mu_t(\boldsymbol{\theta})}{\sqrt{v_t(\boldsymbol{\theta}) + \sigma_n^2}}\right) & \text{if } f(\boldsymbol{\theta}) = \log \Delta_{\boldsymbol{\theta}}, \end{cases} \tag{47}$$

where $h$ is the bandwidth (threshold). The function $F(x)$ was defined in Equation (28) and denotes the cdf of a standard normal random variable, and $\mu_t$ and $v_t + \sigma_n^2$ are the posterior mean and variance of the Gaussian process.

Both $\hat{L}_u^{(t)}$ in the nonparametric approach and $\hat{L}_s^{(t)} = \exp(\hat{\ell}_s^{(t)})$ in the parametric approach are computable approximations $\hat{L}$ of the likelihood $L$. Evaluating them is cheap since no further runs of the simulator are needed. Derivatives can also be computed since the derivatives of the posterior mean and variance are tractable for Gaussian processes. A given approximate likelihood function can thus be used in various ways for inference: We can maximize it and compute its curvature (Hessian matrix) to obtain error bars, we can perform inference with a hybrid Monte Carlo algorithm in a MCMC framework, or use it according to Equation (11) in an importance sampling approach.

For the results in this paper, we used iterative importance sampling where in each iteration, the auxiliary pdf $q$ was a mixture of Gaussians as in Section 3.5. The initial auxiliary pdf was defined as a mixture of Gaussians in the same manner by associating uniform weights with the $\boldsymbol{\theta}^{(i)}$ acquired in the Bayesian optimization step. Samples from the prior pdf $p_{\boldsymbol{\theta}}$ are not needed in such an approach, which can be advantageous if obtaining them is expensive.

We next illustrate model-based likelihood approximation using the example about bacterial infections in day care centers.

**Example 11** (Model-based approximate univariate likelihoods for the day care centers)**.** We inferred the likelihood function via Bayesian optimization using a Gaussian process model with quadratic prior mean and $T = 50$ data points (10 initial points and 40 acquisitions). The bandwidths and general setup were as in Example 7. The left column of Figure 8 shows the estimated models of the discrepancies for the different parameters and compares them with the empirical distributions reported in Figure 5. The right column of Figure 8 shows the estimated likelihood functions $\hat{L}_u^{(t)}$, $t = 50$ (blue solid curves), and compares them with the sample average based approximations $\hat{L}_u^N$ from Figure 5 (black, dots). For Bayesian optimization, the computational cost for an entire likelihood curve was 50 simulations. This is in stark contrast to the computational cost of $N = 300$ simulations for a single evaluation of $\hat{L}_u^N$ in the sample-based approach. Since $\hat{L}_u^N$ was evaluated on a grid of 50 points, the model-based results required 300 times fewer simulations. The computational savings were achieved through the use of smooth regression and the active construction of the training data in Bayesian optimization. ▲

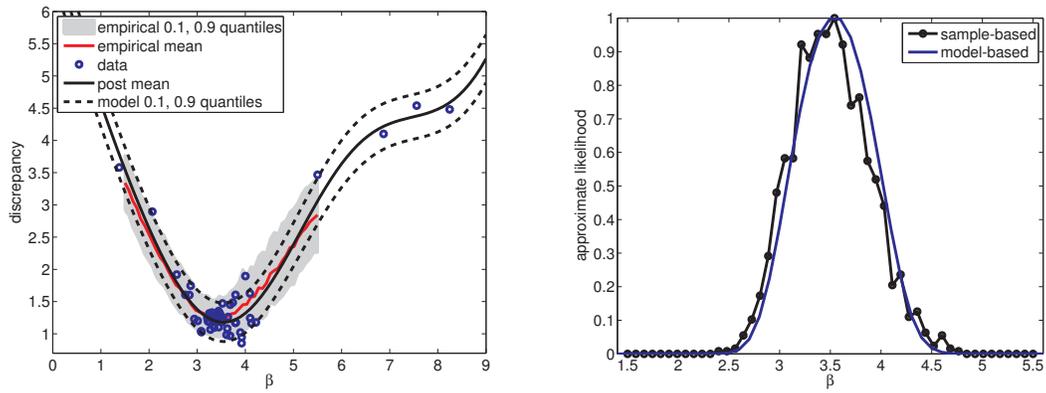

(a) Internal infection parameter $\beta$

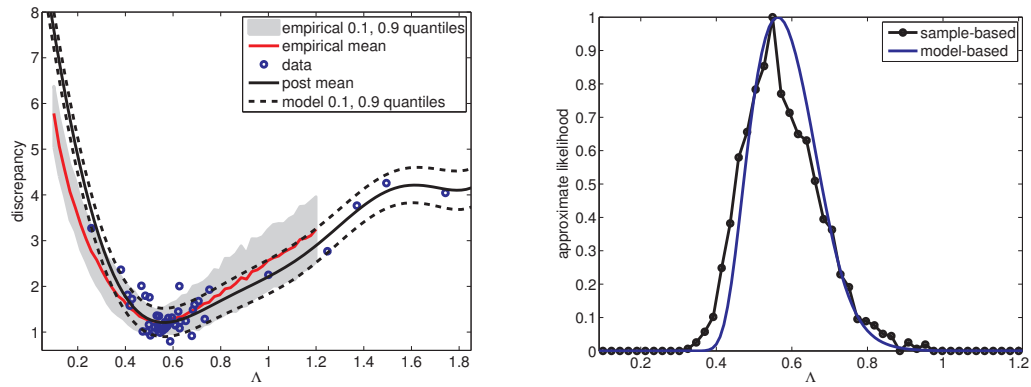

(b) External infection parameter $\Lambda$

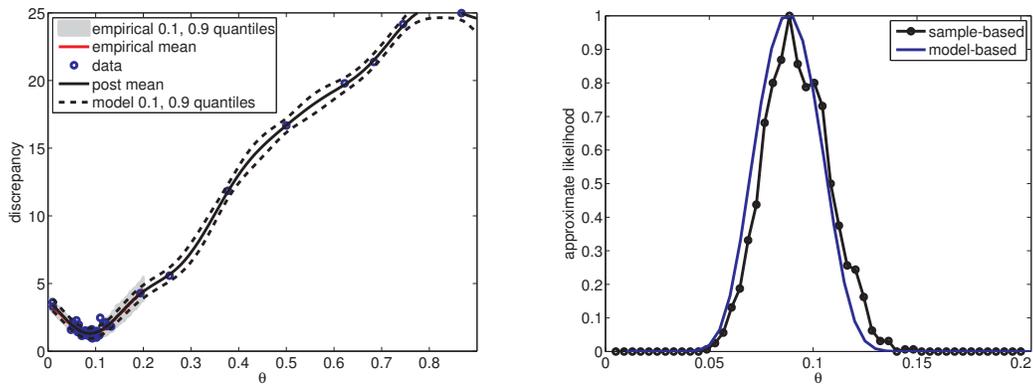

(c) Co-infection parameter $\theta$

Figure 8: Distributions of the discrepancies and approximate likelihoods for the day care center example. For reference, the sample average results are reproduced from Figure 5. In the standard sample average approach, each likelihood curve required 15,000 simulations (right column, black lines with markers). In the proposed model-based approach, each likelihood curve required 50 simulations (right column, blue solid lines). This yields a factor of 300 in computational savings.

### 6.4 Possible Extensions

We use in this paper a basic version of Bayesian optimization to do likelihood-free inference. But more advanced versions exist which opens up a range of possible extensions.

#### 6.4.1 SCALABILITY WITH THE NUMBER OF ACQUISITIONS

The straightforward approach of Section 6.1 to Bayesian optimization with a Gaussian process model requires the inversion of the $t \times t$ matrix $\mathbf{K}_t$. The inversion has complexity $\mathcal{O}(t^3)$ which limits the number of acquisitions to a few thousands. For the applications in this paper, this has not been an issue but we would like to be able to acquire more than a few thousand points if necessary.

Research on Gaussian processes has produced numerous methods to deal with the inversion of $\mathbf{K}_t$ (for an overview, see Rasmussen and Williams, 2006, Chapter 8). Importantly, we can directly use any of these methods for the purpose of likelihood-free inference. For example, sparse Gaussian process regression employs $m < t$ "inducing variables" to reduce the complexity from $\mathcal{O}(t^3)$ to $\mathcal{O}(tm^2)$ (see, for example, Quiñonero-Candela and Rasmussen, 2005). The inducing variables and the hyperparameters of the Gaussian process can be optimized using variational learning (Titsias, 2009), which is also amenable to stochastic optimization to further reduce the computational cost (Hensman et al., 2013).

An alternative approach to Gaussian process regression is Bayesian linear regression with a set of $m < t$ suitably chosen basis functions. The two approaches are closely related (see, for example, Rasmussen and Williams, 2006, Chapter 2), but instead of a $t \times t$ matrix, a $m \times m$ matrix needs to be inverted. This reduces the computational complexity again to $\mathcal{O}(tm^2)$. In order to keep the number of required basis functions small, adaptive basis regression with deep neural networks has been employed to perform Bayesian optimization (Snoek et al., 2015).

#### 6.4.2 HIGH-DIMENSIONAL INFERENCE

Likelihood-free inference is in general very difficult when the dimensionality $d$ of the parameter space is large. This difficulty manifests itself in our approach in the form of a nonlinear regression problem which needs to be solved. While we are only interested in accurate regression results in the areas of the parameter space where the discrepancy is small, discovering these areas becomes more difficult as the dimension increases.

In general, more training data are needed with increasing dimensions so that a method which can handle a large number of acquisitions is likely required (see above). Furthermore, the optimization of acquisition functions is also more difficult in higher dimensions.

Bayesian optimization in high dimensions typically relies on structural assumptions about the objective function. In recent work, it was assumed that the objective varies along a low dimensional subspace only (Chen et al., 2012; Wang et al., 2013; Djolonga et al., 2013), or that it takes the form of an additive model (Kandasamy et al., 2015). This work and further developments in high-dimensional Bayesian optimization can be leveraged for the challenging problem of high-dimensional likelihood-free inference.

### 6.4.3 Parallelization and Acquisition Rules

Bayesian optimization lends itself to parallelization. In particular the acquisition of new data points can be performed in parallel. While several well-known acquisition rules are sequential, they can also be parallelized. Our stochastic acquisition rule provides an easy mechanism by using a sequential rule to define a probability distribution for the location of the next acquisition. Several points can then be drawn in parallel from that distribution. We employ the lower confidence bound selection criterion in Equation (45) to drive the stochastic acquisitions, but alternative rules, for example the maximization of expected improvement, can be used in an analogous way. Other stochastic acquisition rules, like for instance Thompson sampling (Thompson, 1933; Chapelle and Li, 2011; Russo and Van Roy, 2014), enable similarly the concurrent acquisition of multiple data points.

A more elaborate way to parallelize a sequential acquisition rule is to design the joint acquisitions such that the resulting algorithm behaves as if the points are chosen sequentially (Azimi et al., 2010), or to integrate out the possible outcomes of the pending function evaluations (Snoek et al., 2012). Moreover, parallel versions of the lower/upper confidence bound criterion have been proposed by Contal et al. (2013) and Desautels et al. (2014).

In most theoretical studies on acquisition rules, the objective function in Bayesian optimization is modeled as a Gaussian process with uncorrelated Gaussian observation noise. The distribution of the (log) discrepancy, however, may not follow this assumption. This implies on the one hand that the probabilistic modeling of the discrepancy could be improved (see below). On the other hand, it also means that further research would be needed about optimal acquisition rules in the context of likelihood-free inference.

### 6.4.4 Probabilistic Model

We modeled the discrepancy $\Delta_{\boldsymbol{\theta}}$ as a Gaussian or log Gaussian process using a squared exponential covariance function and uncorrelated Gaussian observation noise. While simple and often used, we are not limited to these choices. The literature on Gaussian process regression and Bayesian optimization provides several alternatives and extensions (for an overview, see Rasmussen and Williams, 2006). Modeling of $\Delta_{\boldsymbol{\theta}}$ is important because the model affects the inferences made.

In the employed model, a stationary prior distribution is assumed. However, depending on the simulator, the discrepancy may behave differently in different parameter regions. In particular its variance may be input dependent (heteroscedasticity). Such cases can be handled by non-stationary covariance functions or by using different stationary processes in different regions of the parameter space (see, for example, Rasmussen and Williams, 2006, Chapters 6 and 9).

Equation (42) shows that for the Gaussian process model, the posterior variance does not depend on the observed function values but only on the acquisition locations. As more points are acquired in a neighborhood of a point, the posterior variance may shrink even if the observed function values have a larger than expected spread. A dependency on the observed values can be obtained indirectly by updating the hyperparameters of the covariance function. But a more direct dependency may be preferable. An option is to use Student's $t$ processes instead where the posterior variance depends on the observed function values through a global scaling factor (Shah et al., 2014).

## 7. Applications

We here apply the developed methodology to infer the complete Ricker model and the complete model of bacterial infections in day care centers. As in the previous section, using Bayesian optimization in likelihood-free inference (BOLFI) reduces the amount of required simulations by several orders of magnitude.

### 7.1 Ricker Model

We introduced the Ricker model in Example 2. It has three parameters: $\log r$, $\sigma$, and $\phi$. The difficulty in the inference stems from the dynamics of the latent time series and the unobserved variables. We inferred the parameters using the synthetic likelihood of Wood (2010) from the data shown in Figure 1(a) which were generated with $\boldsymbol{\theta}_o = (3.8, 0.3, 10)$.

Wood (2010) inferred the model with a random walk Markov chain Monte Carlo algorithm using $\hat{\ell}_s^N(\boldsymbol{\theta})$ with $N = 500$. The random walk was defined on the log-parameters due to their positivity. In a baseline study with the computer code made publicly available by Wood (2010), we were not able to infer the parameters with the settings in the original publication (Wood, 2010, Section 1.1 in the supplementary material). Reducing the proposal standard deviation for $\sigma$ by a factor of ten enabled inference even though different Markov chains still led to rather different marginal posterior pdfs for $\sigma$. These issues were observed for $N \in \{500, 1000, 5000\}$ and for Markov chains run twice as long as in the original publication (100,000 versus 50,000 iterations). In addition to the usual random effects in MCMC, the variability in the outcomes of the different chains may be due to his approach of working on a single realization of the random log synthetic likelihood function (see Figure 3 for example realizations when only $\log r$ is varied). The results of our baseline study are reported in Appendix D. Given the nature of the baseline results, we should not expect that the results from our method match them exactly.

For BOLFI, we modeled the random log synthetic likelihood $\hat{\ell}_s^N$ as a log-Gaussian process with a quadratic prior mean function (using $N = 500$ as Wood, 2010). Bayesian optimization was performed with the stochastic acquisition rule and 20 initial data points. Figure 9 shows $-\hat{\ell}_s^{(t)}$ for $t \in \{50, 150, 500\}$. The results for $t = 50$ and $t = 500$ differ more in the shape of the estimated regression functions than in the location of the optima. As the evidence set grows, the algorithm learns that the log synthetic likelihood is less confined along $\sigma$ and that the curvature along the other dimensions should be larger. The plot also shows that there is a negative correlation between $\log r$ and $\phi$ (conditional on $\sigma$). This is reasonable since a larger growth rate $r$ can be compensated with a smaller value of the observation scalar $\phi$ and vice versa.

The approximation $\hat{\ell}_s^{(t)}$ was used to perform posterior inference of the parameters via the iterative importance sampling scheme of Section 6.3 (using three iterations with 25,000 samples each). This sampling is purely model-based and does not require further runs of the simulator. The computed marginal posterior pdfs are shown in Figure 10 (curves in gray) together with a MCMC solution for reference (blue dashed). It can be seen that already after $t = 150$ acquired data points, we obtain a solution which matches the MCMC solution well at a fraction of the computational cost. About 600 times fewer calls to the simulator were needed.

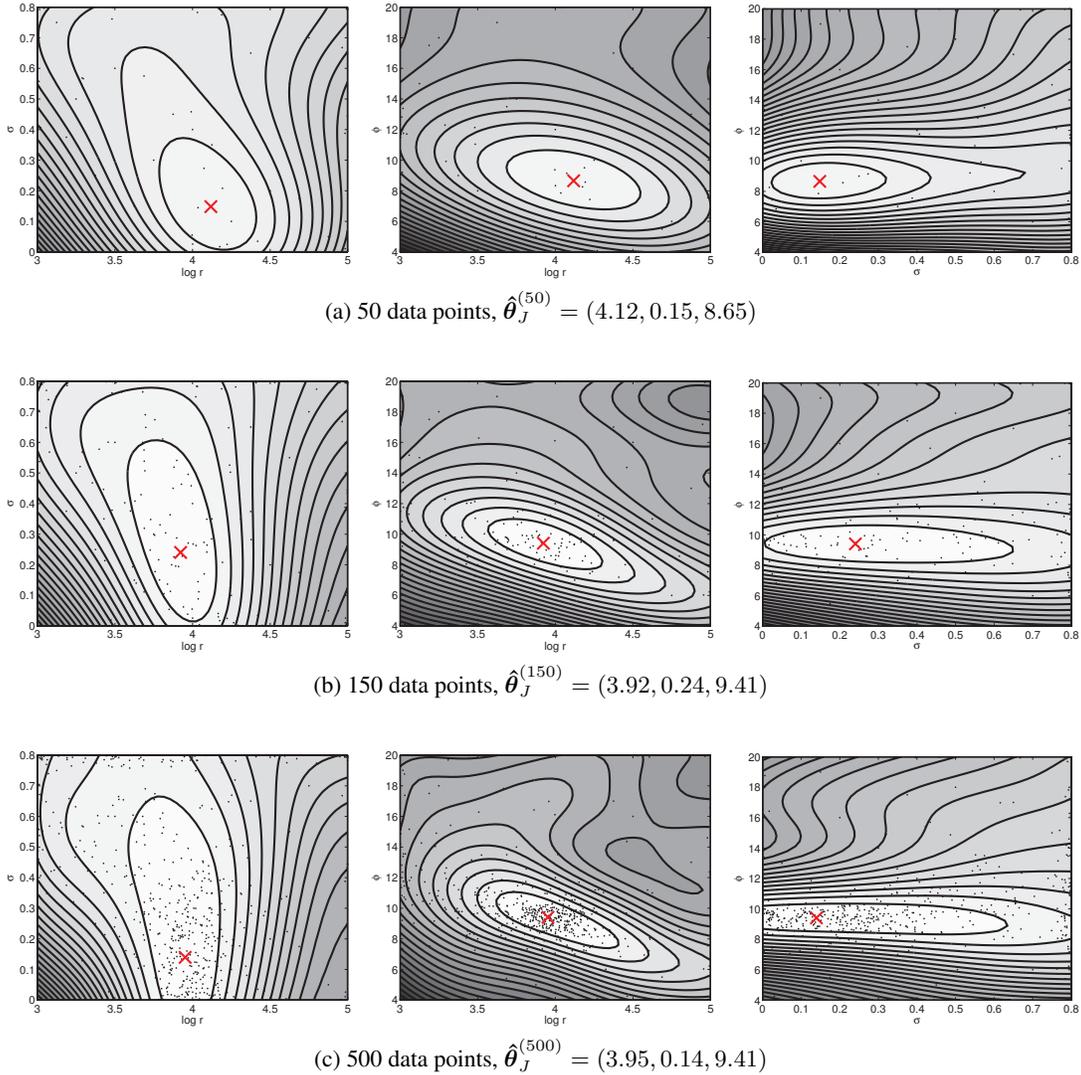

(a) 50 data points, $\hat{\boldsymbol{\theta}}_J^{(50)} = (4.12, 0.15, 8.65)$

(b) 150 data points, $\hat{\boldsymbol{\theta}}_J^{(150)} = (3.92, 0.24, 9.41)$

(c) 500 data points, $\hat{\boldsymbol{\theta}}_J^{(500)} = (3.95, 0.14, 9.41)$

Figure 9: Isocontours of the estimated negative log synthetic likelihood function for the Ricker model. Each panel shows slices of $-\hat{\ell}_s^{(t)}$ with $\operatorname{argmax} \hat{\ell}_s^{(t)}$ as center point when two of the three variables are varied at a time. The center points are marked with a red cross. The dots mark the location of the acquired parameters $\boldsymbol{\theta}^{(i)}$ (projected onto the plane). The intensity map is the same in all figures; white corresponds to the smallest value.

The largest differences between the model-based and the MCMC solution occur for parameter $\sigma$ (Figure 10(b)). But we have seen that this is a difficult parameter to infer and that the MCMC solution may actually not correspond to ground truth. The two posteriors inferred by MCMC have, for instance, posterior means (blue diamonds) which are further from the data generating parameter $\sigma_o = 0.3$ (green circle) than our model-based solution (black square). For the other parameters, the posterior means of the model-based solution are also closer to ground truth than the posterior means of the MCMC solution.

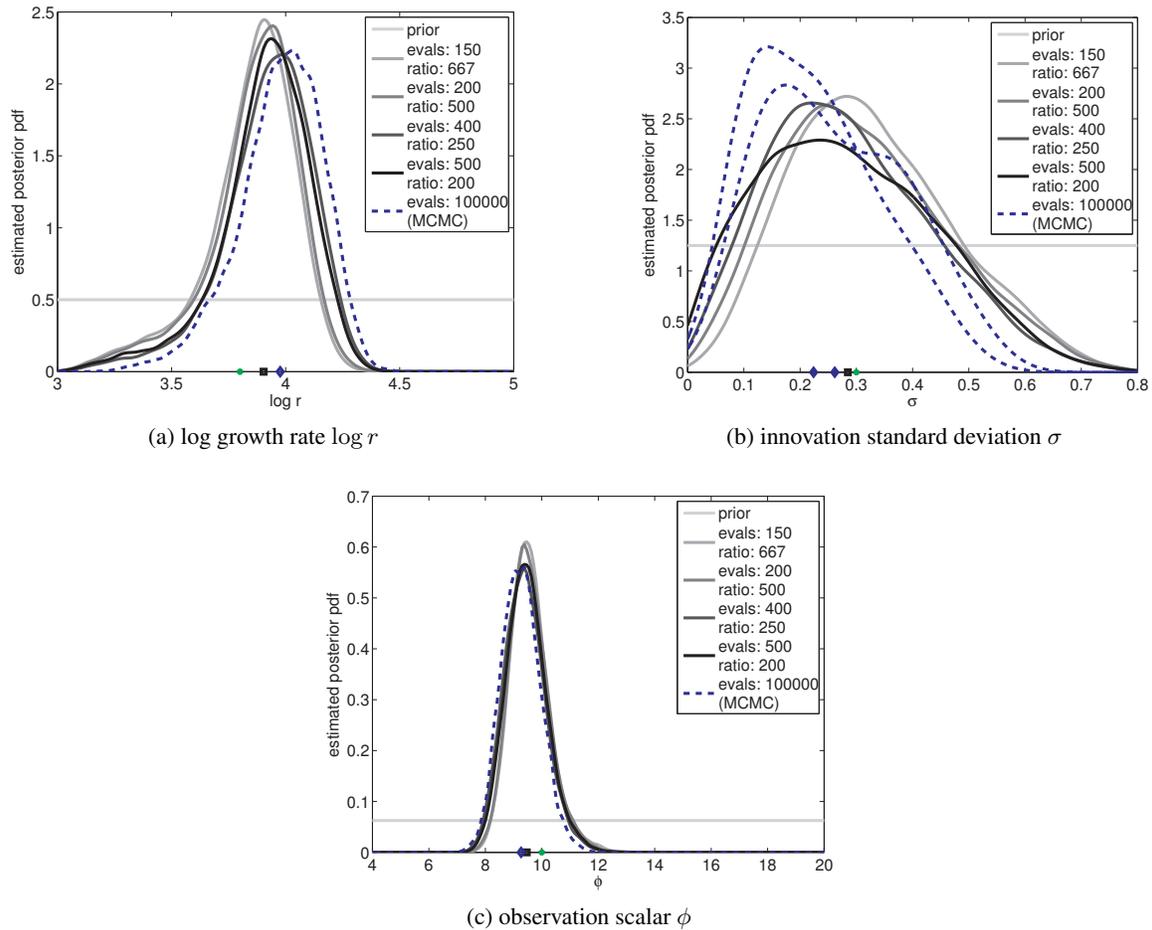

(a) log growth rate log $r$

(b) innovation standard deviation $\sigma$

(c) observation scalar $\phi$

Figure 10: Marginal posterior pdfs for the Ricker model. The model-based solutions are shown in gray, the blue dashed curves are the MCMC solution. The green circles on the x-axes mark the location of $\boldsymbol{\theta}_o = (3.8, 0.3, 10)$. The blue diamonds mark the value of the posterior mean for the MCMC solution while the black squares indicate the posterior means of the model-based solution. For the MCMC solution, 100,000 simulated data sets are needed. Bayesian optimization yields informative solutions using 150 simulated data sets only, which corresponds to 667 times fewer simulations than with MCMC.

### 7.2 Bacterial Infections in Day Care Centers

The model for bacterial infections in day care centers was described in Example 3. It has three parameters of interest: $\beta$, $\Lambda$, and $\theta$. The likelihood function is intractable due to the infinitely many unobserved correlated variables. We inferred the model using the discrepancy $\Delta_{\boldsymbol{\theta}}$ described in Example 7 from the same real data as Numminen et al. (2013).

For BOLFI, we modeled the discrepancy $\Delta_{\boldsymbol{\theta}}$ as a Gaussian process with a quadratic prior mean function and used the stochastic acquisition rule. The algorithm was initialized with 20 data points. Figure 11 shows the estimated regression functions $\hat{J}^{(t)}$ for $t \in \{50, 100, 150, 500\}$. For $t = 50$, the optimal co-infection parameter $\theta$ is at a boundary of the parameter space. As more training data are acquired, the shape of the estimated regression function changes. The algorithm learns that the optimal $\theta$ is located away from the boundary, and the isocontours become oblique which indicates a negative (conditional) correlation between all three parameters. A negative correlation between $\beta$ and $\Lambda$ given the estimate of $\theta$ is reasonable because an increase in transmissions inside the day care centers (increase of $\beta$) can be compensated with a decrease of transmissions from an outside source (decrease of $\Lambda$). The co-infection parameter $\theta$ is negatively correlated with $\beta$ given the estimate of $\Lambda$ because a decrease in the tendency to be infected by multiple strains of the bacterium (decrease of $\theta$) can be offset by an increase of the transmission rate (increase of $\beta$). The same reasoning applies to $\Lambda$ given a fixed value of $\beta$.

We used the Gaussian process model of the discrepancy to compute the model-based likelihood $\hat{L}_u^{(t)}$. The threshold $h$ was chosen as the 0.05 quantile of the modeled discrepancy at the minimizer of the estimated regression function. Model-based posterior inference was then performed via iterative importance sampling as described in Section 6.3 (using three iterations with 25,000 samples each). Figure 12 (left column) shows the inferred marginal posterior pdfs. They stabilize quickly as the amount of acquired data increases.

The right column in Figure 12 compares our model-based results with the solution by Numminen et al. (2013) (blue horizontal lines with triangles) and with results by the population Monte Carlo (PMC) ABC algorithm of Section 3.5 (black curves with diamonds). Numminen et al. (2013) used a PMC-ABC algorithm as well but with a slightly different discrepancy measure (see Example 7). Both PMC results were obtained using 10,000 initial simulations to set the initial threshold, followed by four more iterations with shrinking thresholds where in each iteration, data sets were simulated till 10,000 accepted parameters were obtained. It can be seen that the posterior mean and the credibility intervals of the two PMC results match in the fourth generation, which indicates that our modification of the discrepancy measurement had a negligible influence. For the PCM results shown in black, iteration one to four required 121,374; 277,997; 572,007; and 1,218,382 simulations each, giving a total computational cost of 2,199,760 simulations for the results of iteration four. In terms of computing time, the PMC computations took about 4.5 days on a cluster with 200 cores. Our model-based results for $t = 1,000$ were obtained with one tenth of the initial simulations of the reference methods and took only about 1.5 hours on a desktop computer.[4] Out of the computing time, 93% were spent on simulating the day care centers, and 7% on regression and optimization of the acquisition function.

The posterior means of our model-based approach match quickly the reference results (red curves with circles versus blue curves with triangles). The focus on the modal region yields, however, broader credibility intervals. The broader model-based posterior pdfs suggest that they could be used as auxiliary pdf for PMC-ABC or other iterative ABC algorithms which are based on importance sampling. Moreover, one could evaluate the discrepancy at the sampled points to obtain additional training data in order to refine the model.

---

4. The simulation of the 29 day care centers in the model was partly parallelized by means of seven cores.

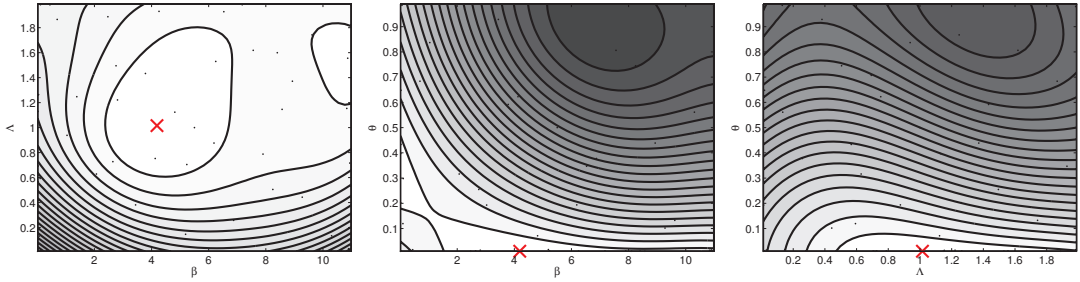

(a) 50 data points, $\hat{\boldsymbol{\theta}}_J^{(50)} = (4.20, 1.02, 0.01)$

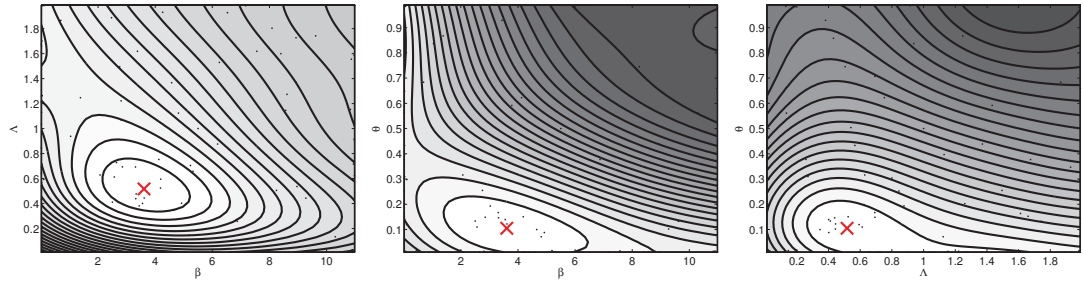

(b) 100 data points, $\hat{\boldsymbol{\theta}}_J^{(100)} = (3.61, 0.52, 0.11)$

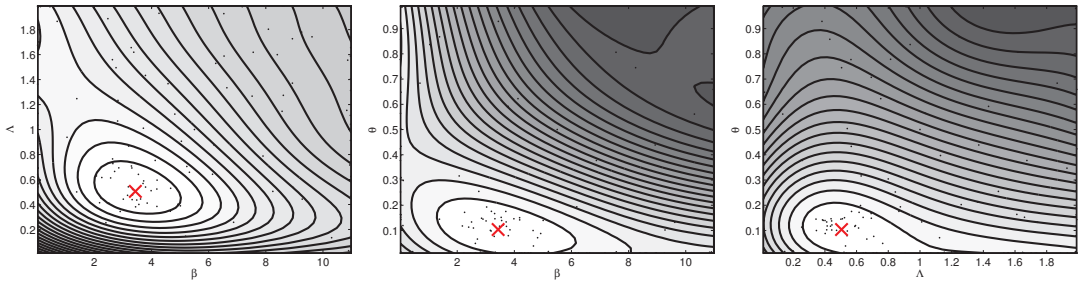

(c) 150 data points, $\hat{\boldsymbol{\theta}}_J^{(150)} = (3.43, 0.50, 0.10)$

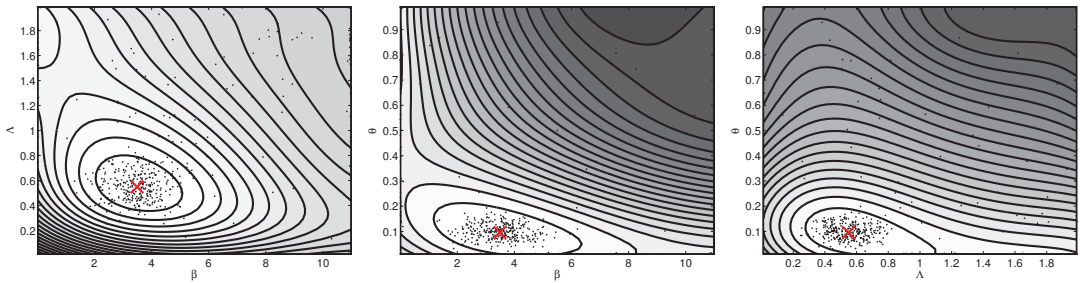

(d) 500 data points, $\hat{\boldsymbol{\theta}}_J^{(500)} = (3.50, 0.55, 0.10)$

Figure 11: Isocontours of the estimated regression function $\hat{J}^{(t)}$ for the day care center model. Visualization is as in Figure 9.

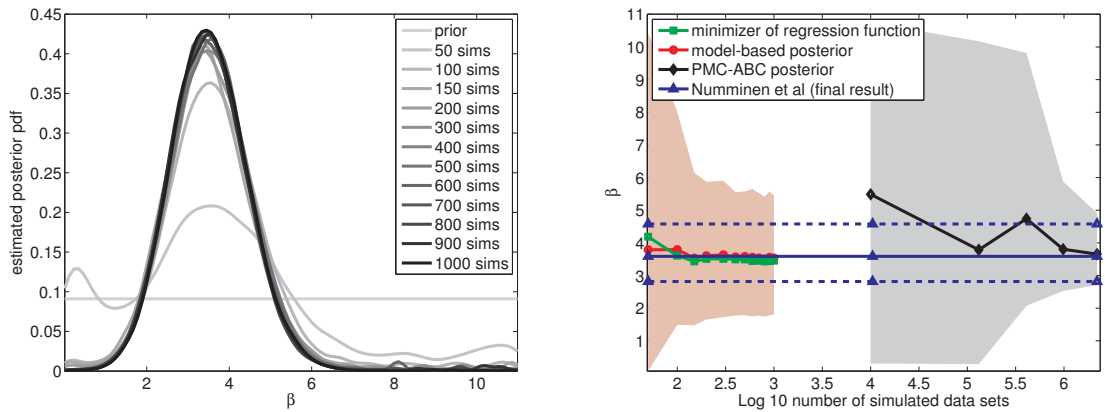

(a) Internal infection parameter $\beta$

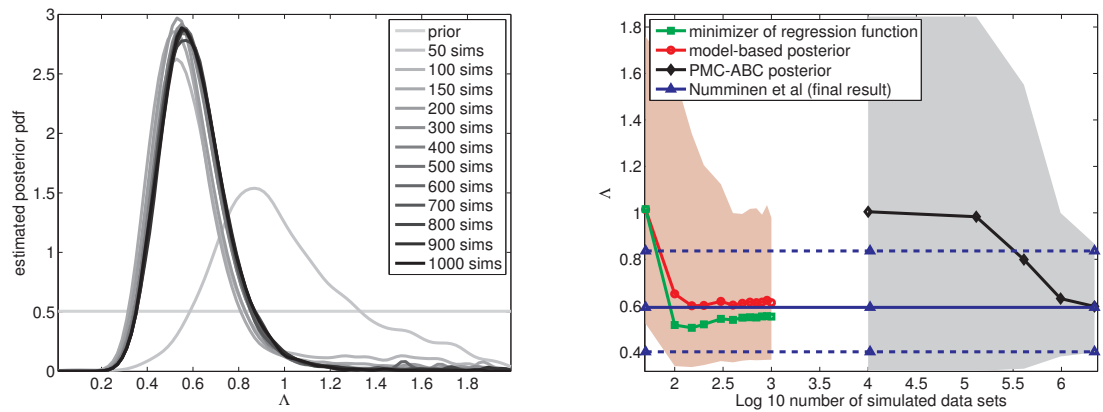

(b) External infection parameter $\Lambda$

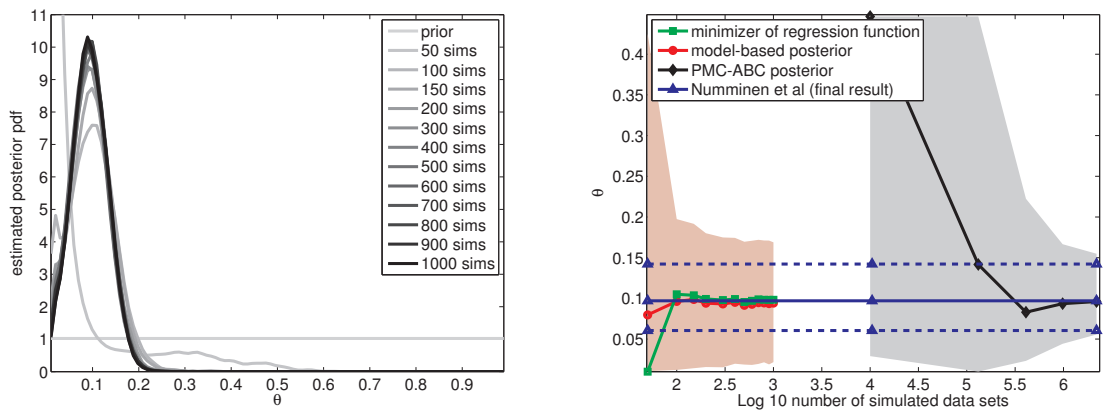

(c) Co-infection parameter $\theta$

Figure 12: Marginal posterior results for the day care center model. The left column shows the obtained model-based posterior pdfs. The right column compares the posterior mean and the 95% credibility interval with results from PMC-ABC algorithms. We obtain conservative estimates of the model parameters at a fraction of the computational cost. Posterior means are shown as solid lines with markers, credibility intervals as shaded areas or dashed lines.

## 8. Conclusions

Our paper dealt with inferring the parameters of simulator-based (generative) statistical models. Inference is difficult for such models because of the intractability of the likelihood function. While it is an open question whether variational principles are also applicable, the parameters of simulator-based statistical models are typically inferred by finding values for which the discrepancy between simulated and observed data tends to be small. We have seen that such an approach is computationally costly. The high cost is largely due to a lack of knowledge about the functional relation between the model parameters and the discrepancies. We proposed to use regression to infer the relation using training data which are actively acquired. The acquisition is performed such that the focus in the regression is on regions in the parameter space where the discrepancy tends to be small. We implemented the proposed strategy using Bayesian optimization where the discrepancy is modeled with a Gaussian process. The posterior distribution of the Gaussian process was used to construct a model-based approximation of the intractable likelihood. This combination of probabilistic modeling and optimization reduced the number of simulated data sets by several orders of magnitude in our applications. The reduction in the number of required simulations accelerated the inference substantially.

Our approach is related to the work by Rasmussen (2003) and the two recent papers by Wilkinson (2014) and Meeds and Welling (2014) (which became only available after we first proposed our approach at "ABC in Rome" in 2013): Rasmussen (2003) used a Gaussian process to model the logarithm of the target pdf in a hybrid Monte Carlo algorithm. There are two main differences to our work. First, a scenario was considered where the target can be evaluated exactly at a finite computational cost, even though the cost might be high. In our case, exact evaluation of the likelihood function is not assumed possible at finite cost. This difference is important because approximate likelihood evaluations might be rather noisy. The second difference is that we used Bayesian optimization to focus on the modal areas of the target.

Related to the approach of Rasmussen (2003), Wilkinson (2014) modeled the log likelihood as a Gaussian process. This is different from our work where we model the discrepancies. We believe that modeling the discrepancies is advantageous because it allows to delay the selection of the kernel and bandwidth which are needed in the nonparametric setting. This is important because it enables one to make use of all simulated data. In the parametric setting, the two modeling strategies lead to identical solutions. We found further that accurate point estimates can be obtained by modeling the discrepancies only. In particular, minimizing their regression function corresponds to maximizing a lower bound of the approximate nonparametric likelihood under mild conditions. As a second difference, Wilkinson (2014) used space-filling points together with a plausibility criterion to obtain the parameter values for the regression. This is in contrast to Bayesian optimization where powerful optimization methods are employed to quickly identify the areas of interest.

Meeds and Welling (2014) proposed an alternative to the sample average approximation of the (limiting) synthetic likelihood by modeling each element of the intractable mean and covariance matrix of the summary statistics with a Gaussian process. The resulting likelihood approximation was used together with a Markov chain Monte Carlo algorithm for posterior inference. The differences to our approach lie in the quantities modeled and in the use Bayesian optimization to actively design the training data for the Gaussian processes.

There are also connections to the body of work on Bayesian analysis of computer codes (for an introduction to this field of research, see for example the paper by O'Hagan, 2006): Sacks et al. (1989) and Currin et al. (1991) modeled the outputs of general deterministic computer codes as Gaussian processes. The computer codes were, for example, solving complex partial differential equations, and the papers were about finding an emulator for the heavy computations. Inference of unknown parameters of the computer codes given observed data was only considered later by Cox et al. (2001) and Kennedy and O'Hagan (2001). The observed and simulated data were modeled using Gaussian processes, and space-filling points were used to choose the parameters for which the computer code was run. The main differences to our approach are again the quantities modeled, and the use of Bayesian optimization.

We employed a rather basic algorithm to perform Bayesian optimization. This does, however, not mean that Bayesian optimization for likelihood-free inference is limited to that particular algorithm. We discussed a number of alternatives, as well as more advanced algorithms which could be used instead, and outlined a general framework for increasing the computational efficiency of likelihood-free inference.

Our paper opens up a wide range of extensions and opportunities for future research. One possibility is to use the tools provided by Bayesian optimization to tackle the challenging problem of likelihood-free inference in high dimensions. More foundational research topics would revolve around the modeling of the discrepancies and the development of acquisition rules which are tailored to the problem of likelihood-free inference. We focused on approximating the modal areas of the intractable likelihoods more accurately than the tails. It is an open question of how to best increase the accuracy in the tail areas. One possibility is to use the samples from the approximate posterior to update the training data for the regression, which would naturally lead to a recursion where the current method would only provide the initial approximation.

## Acknowledgments

This work was partially supported by ERC grant no. 239784 and the Academy of Finland (Finnish Centre of Excellence in Computational Inference Research COIN, 251170). MUG thanks Paul Blomstedt for helpful comments on an early draft of the paper.

Author contributions: MUG proposed, designed, and performed research, and wrote the paper; JC contributed to research design and writing.

## Appendix A. Proof of Proposition 1

We split the objective $\hat{J}_g^N$, defined in Equation (32), into two terms,

$$\hat{J}_g^N(\boldsymbol{\theta}) = T_1(\boldsymbol{\theta}) + T_2(\boldsymbol{\theta}), \tag{48}$$

$$T_1(\boldsymbol{\theta}) = \log|\det \mathbf{C}_{\boldsymbol{\theta}}|, \tag{49}$$

$$T_2(\boldsymbol{\theta}) = \mathrm{E}^N\left[(\Phi_o - \Phi_{\boldsymbol{\theta}})^\top \mathbf{C}_{\boldsymbol{\theta}}^{-1}(\Phi_o - \Phi_{\boldsymbol{\theta}})\right]. \tag{50}$$

Term $T_2$ can be rewritten using the empirical mean $\hat{\boldsymbol{\mu}}_{\boldsymbol{\theta}}$ and the covariance matrix $\hat{\boldsymbol{\Sigma}}_{\boldsymbol{\theta}}$ in Equation (14),

$$T_2(\boldsymbol{\theta}) = \mathrm{E}^N\left[(\Phi_o - \hat{\boldsymbol{\mu}}_{\boldsymbol{\theta}} + \hat{\boldsymbol{\mu}}_{\boldsymbol{\theta}} - \Phi_{\boldsymbol{\theta}})^\top \mathbf{C}_{\boldsymbol{\theta}}^{-1}(\Phi_o - \hat{\boldsymbol{\mu}}_{\boldsymbol{\theta}} + \hat{\boldsymbol{\mu}}_{\boldsymbol{\theta}} - \Phi_{\boldsymbol{\theta}})\right] \tag{51}$$

$$= \mathrm{E}^N\left[(\Phi_o - \hat{\boldsymbol{\mu}}_{\boldsymbol{\theta}})^\top \mathbf{C}_{\boldsymbol{\theta}}^{-1}(\Phi_o - \hat{\boldsymbol{\mu}}_{\boldsymbol{\theta}}) + (\hat{\boldsymbol{\mu}}_{\boldsymbol{\theta}} - \Phi_{\boldsymbol{\theta}})^\top \mathbf{C}_{\boldsymbol{\theta}}^{-1}(\hat{\boldsymbol{\mu}}_{\boldsymbol{\theta}} - \Phi_{\boldsymbol{\theta}})\right.$$
$$\left. + 2(\Phi_o - \hat{\boldsymbol{\mu}}_{\boldsymbol{\theta}})^\top \mathbf{C}_{\boldsymbol{\theta}}^{-1}(\hat{\boldsymbol{\mu}}_{\boldsymbol{\theta}} - \Phi_{\boldsymbol{\theta}})\right] \tag{52}$$

$$= (\Phi_o - \hat{\boldsymbol{\mu}}_{\boldsymbol{\theta}})^\top \mathbf{C}_{\boldsymbol{\theta}}^{-1}(\Phi_o - \hat{\boldsymbol{\mu}}_{\boldsymbol{\theta}}) + \mathrm{tr}\left(\mathbf{C}_{\boldsymbol{\theta}}^{-1} \mathrm{E}^N\left[(\hat{\boldsymbol{\mu}}_{\boldsymbol{\theta}} - \Phi_{\boldsymbol{\theta}})(\hat{\boldsymbol{\mu}}_{\boldsymbol{\theta}} - \Phi_{\boldsymbol{\theta}})^\top\right]\right) \tag{53}$$

$$= (\Phi_o - \hat{\boldsymbol{\mu}}_{\boldsymbol{\theta}})^\top \mathbf{C}_{\boldsymbol{\theta}}^{-1}(\Phi_o - \hat{\boldsymbol{\mu}}_{\boldsymbol{\theta}}) + \mathrm{tr}\left(\mathbf{C}_{\boldsymbol{\theta}}^{-1} \hat{\boldsymbol{\Sigma}}_{\boldsymbol{\theta}}\right), \tag{54}$$

where we have used that $\mathrm{E}^N[\Phi_{\boldsymbol{\theta}}] = \hat{\boldsymbol{\mu}}_{\boldsymbol{\theta}}$. For $\mathbf{C}_{\boldsymbol{\theta}} = \hat{\boldsymbol{\Sigma}}_{\boldsymbol{\theta}}$, we have

$$T_2(\boldsymbol{\theta}) = (\Phi_o - \hat{\boldsymbol{\mu}}_{\boldsymbol{\theta}})^\top \hat{\boldsymbol{\Sigma}}_{\boldsymbol{\theta}}^{-1}(\Phi_o - \hat{\boldsymbol{\mu}}_{\boldsymbol{\theta}}) + p. \tag{55}$$

Hence, for $\mathbf{C}_{\boldsymbol{\theta}} = \hat{\boldsymbol{\Sigma}}_{\boldsymbol{\theta}}$, $\hat{J}_g^N$ equals

$$\hat{J}_g^N(\boldsymbol{\theta}) = \log|\det \hat{\boldsymbol{\Sigma}}_{\boldsymbol{\theta}}| + (\Phi_o - \hat{\boldsymbol{\mu}}_{\boldsymbol{\theta}})^\top \hat{\boldsymbol{\Sigma}}_{\boldsymbol{\theta}}^{-1}(\Phi_o - \hat{\boldsymbol{\mu}}_{\boldsymbol{\theta}}) + p. \tag{56}$$

On the other hand, the log synthetic likelihood $\hat{\ell}_s^N$ is

$$\hat{\ell}_s^N(\boldsymbol{\theta}) = -\frac{p}{2}\log(2\pi) - \frac{1}{2}\log|\det \hat{\boldsymbol{\Sigma}}_{\boldsymbol{\theta}}| - \frac{1}{2}(\Phi_o - \hat{\boldsymbol{\mu}}_{\boldsymbol{\theta}})^\top \hat{\boldsymbol{\Sigma}}_{\boldsymbol{\theta}}^{-1}(\Phi_o - \hat{\boldsymbol{\mu}}_{\boldsymbol{\theta}}), \tag{57}$$

so that

$$\hat{J}_g^N(\boldsymbol{\theta}) = p - p\log(2\pi) - 2\hat{\ell}_s^N(\boldsymbol{\theta}). \tag{58}$$

The claimed result follows now from Equation (31),

$$\log L_g^N(\boldsymbol{\theta}) \geq -\frac{p}{2} + \hat{\ell}_s^N(\boldsymbol{\theta}). \tag{59}$$

Replacing the empirical average $\mathrm{E}^N$ with the expectation shows that the limiting quantities $\tilde{\ell}_s$ and $J_g(\boldsymbol{\theta})$,

$$J_g(\boldsymbol{\theta}) = \mathrm{E}\left[\Delta_{\boldsymbol{\theta}}^g\right], \tag{60}$$

are related by an analogous result. In more detail,

$$J_g(\boldsymbol{\theta}) = \log|\det \mathbf{C}_{\boldsymbol{\theta}}| + \mathrm{E}\left[(\Phi_o - \Phi_{\boldsymbol{\theta}})^\top \mathbf{C}_{\boldsymbol{\theta}}^{-1}(\Phi_o - \Phi_{\boldsymbol{\theta}})\right] \tag{61}$$

$$= \log|\det \mathbf{C}_{\boldsymbol{\theta}}| + (\Phi_o - \boldsymbol{\mu}_{\boldsymbol{\theta}})^\top \mathbf{C}_{\boldsymbol{\theta}}^{-1}(\Phi_o - \boldsymbol{\mu}_{\boldsymbol{\theta}}) + \mathrm{tr}\left(\mathbf{C}_{\boldsymbol{\theta}}^{-1}\boldsymbol{\Sigma}_{\boldsymbol{\theta}}\right), \tag{62}$$

where we used the same development which led to Equation (54) but with the expectation instead of $E^N$. Hence, for $\mathbf{C_\theta} = \mathbf{\Sigma_\theta}$, we have the analogous result by definition of $\tilde{\ell}_s$ in Equation (13),

$$J_g(\boldsymbol{\theta}) = p - p\log(2\pi) - 2\tilde{\ell}_s(\boldsymbol{\theta}). \tag{63}$$

It follows by definition of $J_g$ that $\tilde{\ell}_s$ can be seen as a regression function where $\boldsymbol{\theta}$ is the vector of covariates and $\Delta_{\boldsymbol{\theta}}^g$ is, up to constants and the sign, the response variable.

## Appendix B. Using the Prior Distribution of the Parameters in Bayesian Optimization

In the main text, we focused on acquiring training data in regions in the parameter space where the discrepancy $\Delta_{\boldsymbol{\theta}}$ tends to be small, which corresponds to the modal regions of the approximate likelihoods. For highly informative priors $p_{\boldsymbol{\theta}}$ with modal regions far away from the peaks of the likelihood such an approach is suboptimal for posterior inference. Since the prior is typically fairly broad and the likelihood peaked, this situation is not usual. But if it happens, it is better to directly acquire the training data in the modal areas of the posterior. For inference via the synthetic likelihood, this can be straightforwardly done by approximating $\tilde{\ell}_s + \log p_{\boldsymbol{\theta}}$. In Bayesian optimization with $\Delta_{\boldsymbol{\theta}}^g$ as the response variable, the posterior mean $\mu_t$ in Equation (43) would then be replaced by $\tilde{\mu}_t(\boldsymbol{\theta}) = \mu_t(\boldsymbol{\theta}) - 2\log p_{\boldsymbol{\theta}}$. For inference via an nonparametric approximation of the likelihood, the same approach may also work but this warrants further investigations because the regression function $J$ provides only a lower bound for the likelihood. We also note that using the prior $p_{\boldsymbol{\theta}}$ can be helpful if it is known that the parameters do not influence the model independently, causing for instance the discrepancy to be nearly constant along certain directions in the parameter space.

Figure 13 illustrates the basic idea using Example 1 and a prior pdf $p_{\boldsymbol{\theta}}$ (blue curve) which has practically no overlap with the true likelihood $L$ (green curve). The results are for Bayesian optimization with 20 deterministic acquisitions and a Gaussian process model with constant mean function.

## Appendix C. Bayesian Optimization with a Deterministic versus a Stochastic Acquisition Rule

Example 10 illustrated log-Gaussian modeling and the stochastic acquisition rule by means of the Ricker model with the log growth rate $\log r$ as only unknown. We here show the differences between stochastic and deterministic acquisitions in greater detail. The results are for a log-Gaussian process model.

Figure 14 shows the estimated regression functions $\hat{J}^{(t)}$ as obtained with a deterministic acquisition rule like in Figure 7(d) for different $t$. The acquired data points are vertically clustered because the acquisition rule often proposed nearly identical parameters. Figure 15 shows $\hat{J}^{(t)}$ obtained with a stochastic acquisition rule as in Figure 7(f). While both methods lead to a satisfactory approximation of the negative log synthetic likelihood around its minimum, the result with the stochastic acquisition rule seems more stable because the acquired training data are spread out more evenly in the interval of interest.

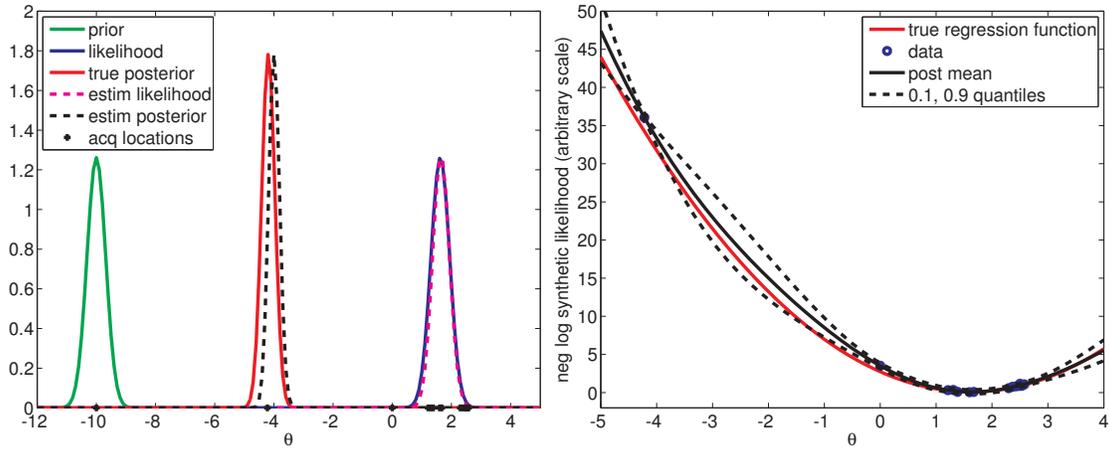

(a) Bayesian optimization without using the prior $p_{\boldsymbol{\theta}}$

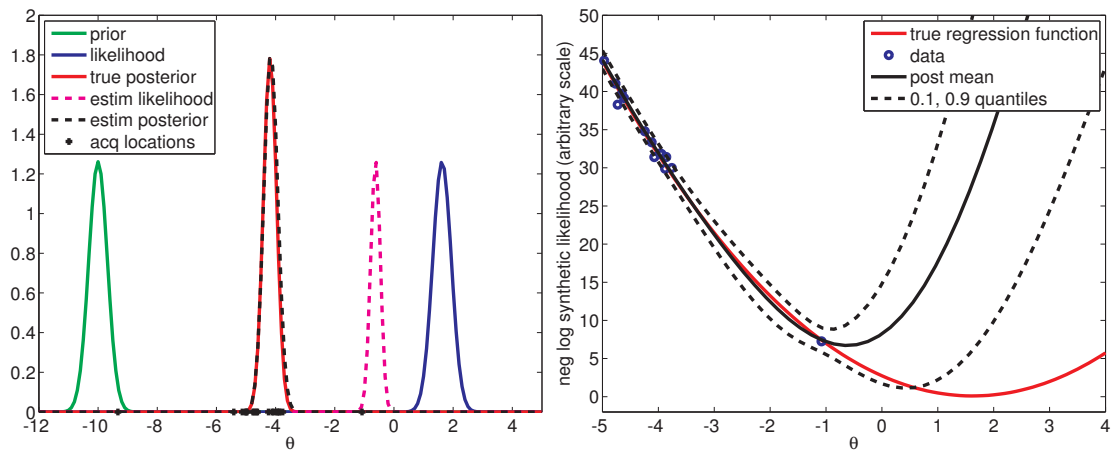

(b) Bayesian optimization with the prior $p_{\boldsymbol{\theta}}$ during the acquisitions

Figure 13: Using the prior density $p_{\boldsymbol{\theta}}$ in Bayesian optimization. (a) If $p_{\boldsymbol{\theta}}$ is not used (or if uniform), the focus is on the modal region of the likelihood. If the prior is far from the mode of the likelihood, the learned model is less accurate in the modal areas of the posterior (black dashed versus red solid curve). (b) The prior pdf $p_{\boldsymbol{\theta}}$ was used to shift the data acquisitions in Bayesian optimization to the modal area of the posterior (see the circles in the figures on the right or on the x-axes on the left). This results in a more accurate approximation of the posterior pdf but a less accurate approximation of the mode of the likelihood (dashed magenta versus blue solid curve).

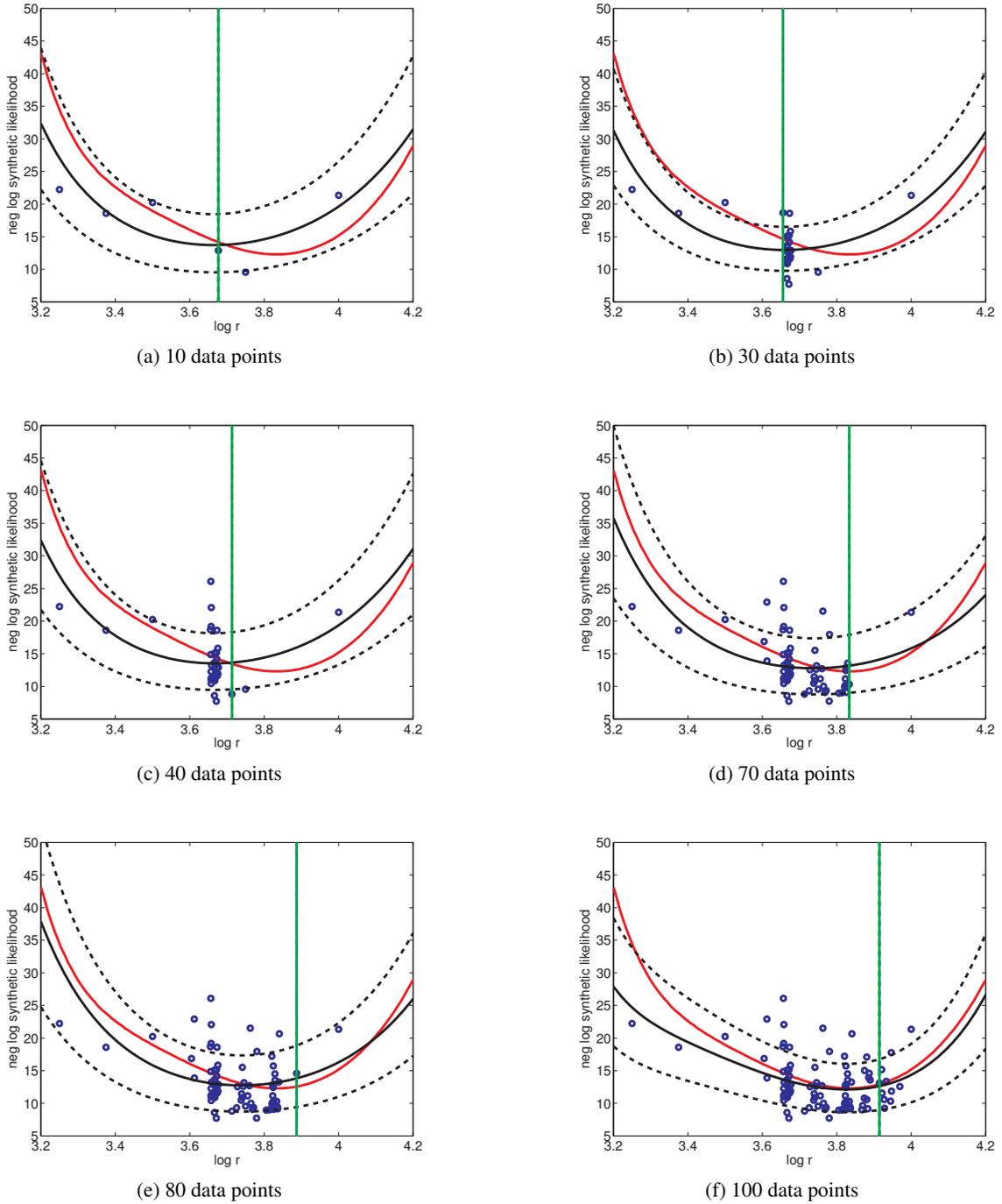

Figure 14: Log-Gaussian process model for the log synthetic likelihood of the Ricker model with $\log r$ as only unknown. The results are for the deterministic acquisition rule consisting of minimization of the acquisition function in Equation (45). Note the vertical clusters. The visualization is as in Figure 7. The plot range was restricted to $(3.2, 4.2)$ so that not all acquisition may be shown.

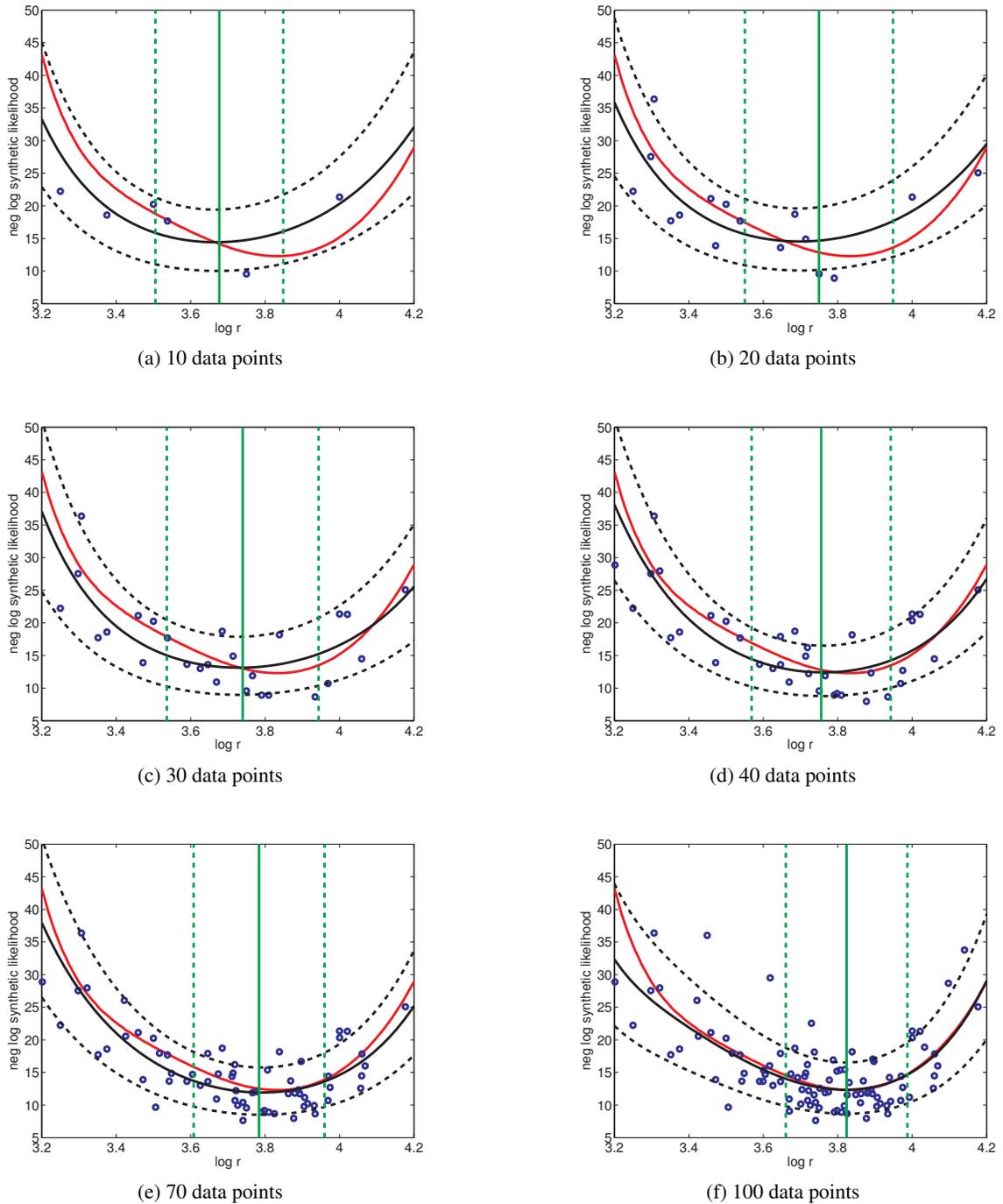

Figure 15: Log-Gaussian process model for the log synthetic likelihood of the Ricker model with $\log r$ as only unknown. The setup and visualization is as in Figure 14 but the stochastic acquisition rule is used. A movie showing the acquisitions and the updating of the model is available at `http://www.cs.helsinki.fi/u/gutmann/material/BOLFI/movies/Ricker1D.avi`.

## Appendix D. Ricker Model Inferred with a Markov Chain Monte Carlo Algorithm

We here report the simulation results for the Ricker model inferred with the log synthetic likelihood $\hat{\ell}_s^N$ and a random walk MCMC algorithm with the code made publicly available by Wood (2010). We ran the algorithm for 100,000 iterations, starting at $\boldsymbol{\theta}_o = (3.8, 0.3, 10)$. The first 25,000 samples were discarded. In the work by Wood (2010), the proposal standard deviation for $\sigma$ was 0.1. Figure 16 shows that this choice led to a chain which got stuck close to $\sigma = 0$ even when $N = 5,000$ (blue, squares). Reducing the proposal standard deviation by a factor of 10 allowed us to obtain reasonable results (red, circles). The proposal standard deviations for the remaining parameters were the same as in the original publication. We then investigated the stability of the inferred posteriors when $N$ is reduced from $N = 5,000$ to $N = 500$ and when the simulator is run with different realizations of the random log synthetic likelihood. Figure 17 shows that the posteriors are stable for $\log r$ and $\phi$ but that there is some variation for $\sigma$.

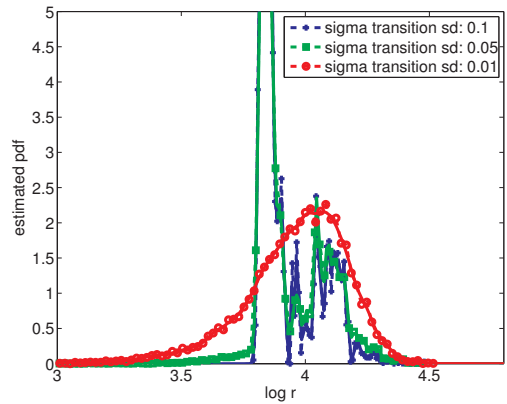

(a) log growth rate $\log r$

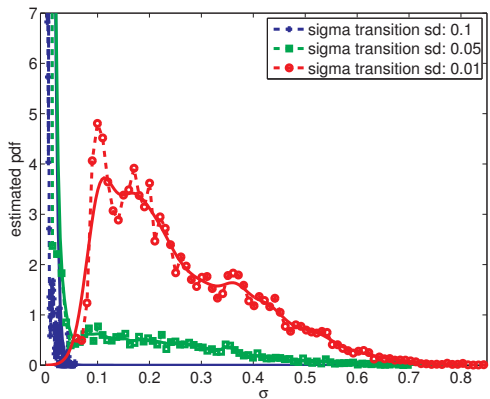

(b) innovation standard deviation $\sigma$

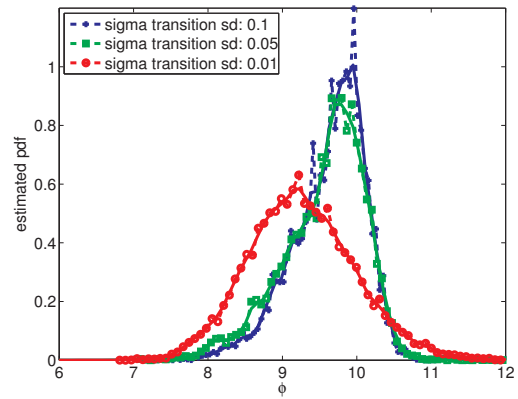

(c) observation scalar $\phi$

Figure 16: Choice of the transition kernels for inference of the Ricker model via MCMC. We used $N = 5,000$ which is ten times more than in the original work (Wood, 2010). The dashed curves with markers are (rescaled) histograms, the solid curves kernel density estimates.

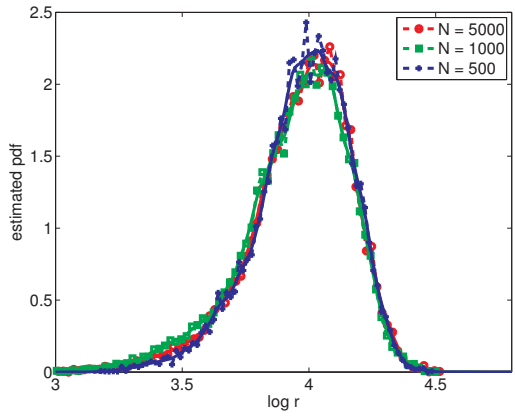

(a) log growth rate $\log r$, varying $N$

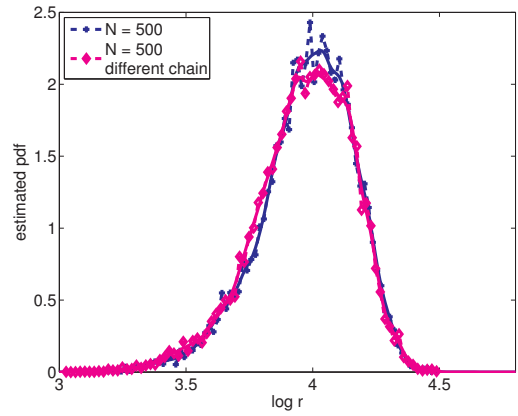

(b) log growth rate $\log r$, varying seed

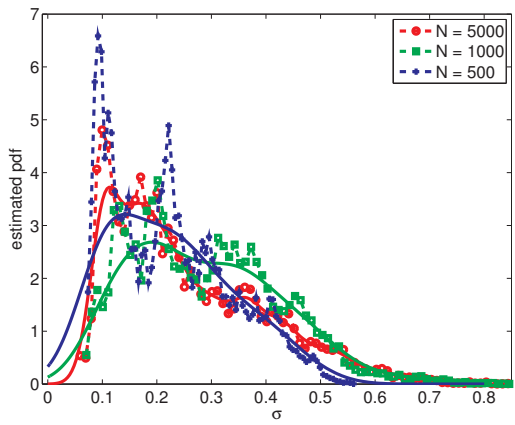

(c) innovation standard deviation $\sigma$, varying $N$

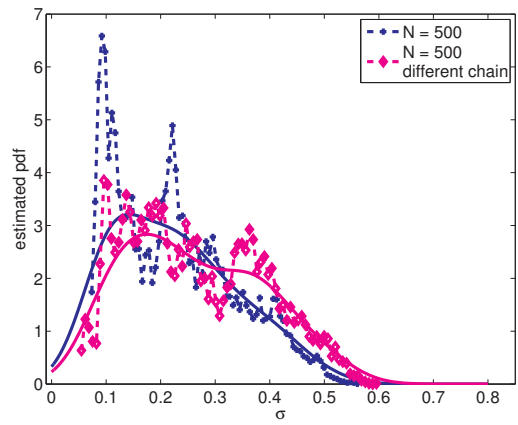

(d) innovation standard deviation $\sigma$, varying seed

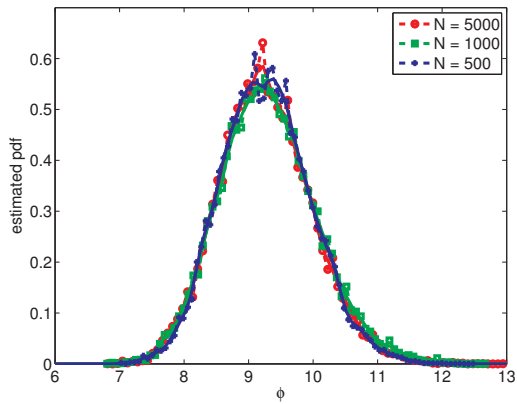

(e) observation scalar $\phi$, varying $N$

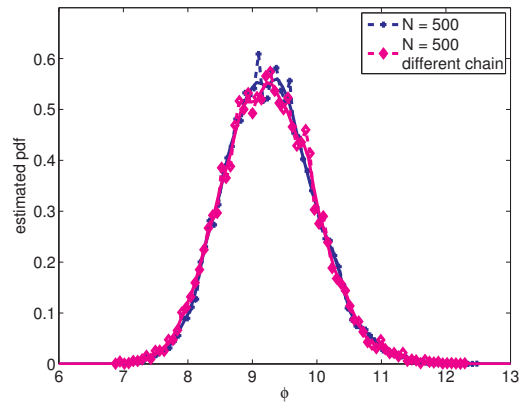

(f) observation scalar $\phi$, varying seed

Figure 17: Effect of the number of simulated data sets $N$ (left column) and the seed of the random number generator (right column) for the Ricker example when inferred with the method by Wood (2010). Visualization is as in Figure 16.